\documentclass[sigconf, nonacm]{acmart}

\makeatletter
\def\@ACM@checkaffil{
    \if@ACM@instpresent\else
    \ClassWarningNoLine{\@classname}{No institution present for an affiliation}%
    \fi
    \if@ACM@citypresent\else
    \ClassWarningNoLine{\@classname}{No city present for an affiliation}%
    \fi
    \if@ACM@countrypresent\else
        \ClassWarningNoLine{\@classname}{No country present for an affiliation}%
    \fi
}
\makeatother

\usepackage{booktabs}
\usepackage{multirow}
\usepackage{array}
\usepackage{tabularx}
\usepackage{makecell}
\usepackage{xcolor}
\usepackage{pifont}
\newcommand{\cmark}{\textcolor{green}{\ding{51}}}
\newcommand{\xmark}{\textcolor{red}{\ding{55}}}

\newcolumntype{Y}{>{\centering\arraybackslash}X}
\usepackage{enumitem}       
\definecolor{deltablue}{rgb}{0.3, 0.55, 0.8}
\newcommand{\up}[1]{\textcolor{deltablue}{\scriptsize +#1}}

\AtBeginDocument{%
  \providecommand\BibTeX{{%
    \normalfont B\kern-0.5em{\scshape i\kern-0.25em b}\kern-0.8em\TeX}}}

\copyrightyear{2026}
\acmYear{2026}

\acmConference[MM '26] {Proceedings of the 34rd ACM International Conference on Multimedia}{November 10-14, 2026}{Rio de Janeiro, Brazil.}
\graphicspath{{./images/}}

\title{JieZi: A Large-Scale Expert-Audited Dataset and Benchmark for Ancient Chinese Character Exegesis}

\author{Ran Li}
\authornote{Both authors contributed equally to this research.}
\affiliation{
  \institution{South China University of Technology}
  \city{Guangzhou}
  \country{China}}
\email{eeran0@mail.scut.edu.cn}

\author{Huiguo He}
\authornotemark[1] 
\affiliation{%
  \institution{South China University of Technology}
  \city{Guangzhou}
  \country{China}}
\email{hehuiguo@scut.edu.cn}

\author{Jiahuan Cao}
\affiliation{%
  \institution{South China University of Technology}
  \city{Guangzhou}
  \country{China}}
\email{eejiahuancao@mail.scut.edu.cn}

\author{Junle Liu}
\affiliation{%
  \institution{South China University of Technology}
  \city{Guangzhou}
  \country{China}}
\email{junle_liu@foxmail.com}

\author{Hiuyi Cheng}
\affiliation{%
  \institution{South China University of Technology}
  \city{Guangzhou}
  \country{China}}
\email{eechenghiuyi1@mail.scut.edu.cn}

\author{Lianwen Jin}
\authornote{Corresponding author.}
\affiliation{%
  \institution{South China University of Technology}
  \city{Guangzhou}
  \country{China}}
\email{eelwjin@scut.edu.cn}

\renewcommand{\shortauthors}{Li, et al.}

\begin{document}


\renewcommand\footnotetextcopyrightpermission[1]{}
\settopmatter{printacmref=false}

\begin{abstract}

The scholarly exegesis of ancient Chinese characters demands integrating visual observation, linguistic analysis, and historical context. However, existing computational approaches focus narrowly on subtasks such as character recognition and retrieval, lacking the structured datasets and benchmarks required for comprehensive scholarly analysis.
To address this limitation, we introduce \textbf{Ancient Chinese Character Exegesis (ACCE)}, a vision-language question answering (VQA) task that models the scholarly exegesis process. ACCE is organized into four progressive levels: basic character identification, glyph-form analysis, meaning exegesis, and diachronic evolution analysis.
To support this task, we construct two complementary resources. 
\textbf{JieZi-Dataset} is the first large-scale, expert-audited VQA training dataset for ACCE, comprising over 500K QA pairs. It is constructed via a pipeline that reduces factual errors by constraining generation with expert-designed templates and source-text references. Human verification is further applied at each key stage to ensure scholarly accuracy.
\textbf{JieZi-Bench} is an evaluation benchmark aligned with the exegesis process, constructed and verified by human experts to ensure evaluation reliability. It consists of four levels with reference answers curated from authoritative lexicographic works held separate from the training data.
Experiments on multimodal large language models show that current models perform well on basic identification but struggle with glyph analysis, semantic reasoning, and diachronic understanding. Fine-tuning on JieZi-Dataset substantially improves performance across all four levels.
Code and dataset are available at https://github.com/Ran00w/JieZi.

\end{abstract}




\maketitle

\begin{figure}[t]
    \centering
    \includegraphics[width=0.95\linewidth]{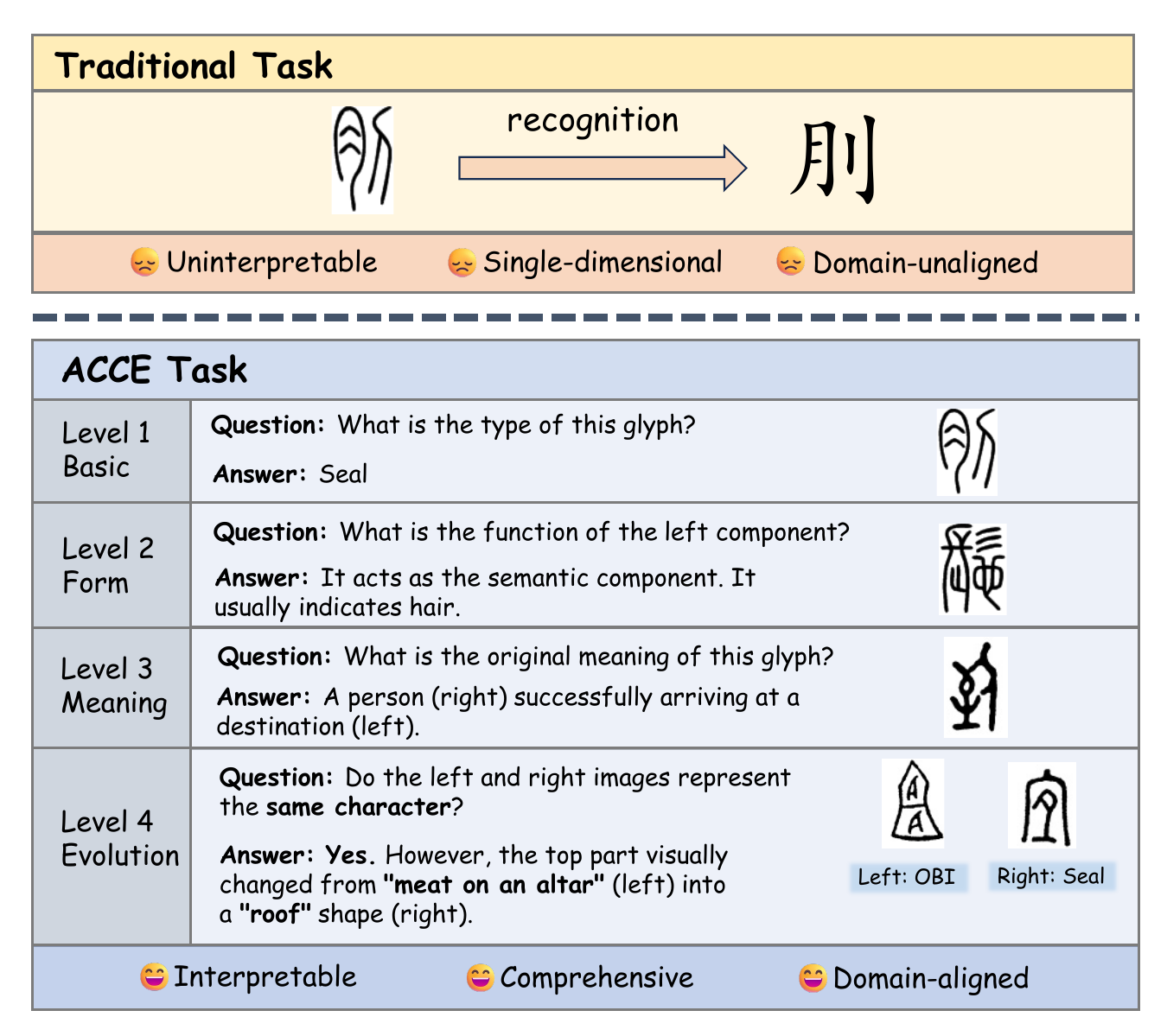}
   \vspace{-10pt}
    \caption{Comparison between the Exegesis task and the traditional Recognition task.}
    \label{fig:intro}
   \vspace{-15pt}
\end{figure}

\section{Introduction}

Ancient Chinese characters represent one of the oldest continuously used writing systems, carrying irreplaceable historical and cultural heritage~\cite{denecke2017oxford, keightley1996art, heinrich2020language, unescoChineseOracleBone}. 
As a core undertaking in ancient Chinese philology, the exegesis of individual ancient glyphs involves analyzing visual form, interpreting semantic function, and tracing diachronic evolution. This process demands years of specialized training, cross-referencing authoritative dictionaries, and reconciling divergent scholarly views~\cite{chapinal2023review, williams2014scribal, smith2017early}. 
It remains labor-intensive, subjective, and difficult to scale when processing the massive volume of unearthed artifacts and manuscripts~\cite{bottero2008shuowen, shen2020phonological, Xing_2011}. 

Artificial intelligence, particularly vision-language modeling, offers a promising path to assist in this analysis~\cite{assael2022restoring, assael2025contextualizing}. However, three fundamental challenges hinder progress toward comprehensive, scholar-grade exegesis.
First, existing works mainly focus on narrow subtasks such as glyph recognition, image retrieval, and single-label classification~\cite{chi-etal-2024-ancient, li2024comprehensivesurveyoraclecharacter, li2024oracle}. These tasks address only fragments of the full exegetical workflow, failing to formalize the whole scholarly exegesis process.
Second, current MLLMs lack domain-specific paleographic knowledge, causing frequent hallucinations when applied to ancient character analysis~\cite{bai2025qwen3vltechnicalreport,kimiteam2026kimik25visualagentic,wang2025internvl35advancingopensourcemultimodal}. Mitigating this deficiency requires large-scale, expert-verified training data; however, the high cost of manual annotation~\cite{sommerschield2023machine, chapinal2023review} and the unreliability of fully automated generation make neither approach alone practicable or scalable.
Third, there is no benchmark that systematically evaluates the full scope of scholarly exegesis, leaving the capabilities of current MLLMs for comprehensive character analysis largely unclear.

To address these challenges, we take the first step toward computational modeling of the scholarly exegesis workflow for ancient Chinese characters. As illustrated in Fig.~\ref{fig:intro}, we formulate \textbf{Ancient Chinese Character Exegesis (ACCE)}, a novel vision-language task that structures the scholarly exegesis process into four progressive levels: \textbf{basic character identification}, \textbf{glyph-form analysis}, \textbf{meaning exegesis}, and \textbf{diachronic evolution analysis}, grounded in established principles of traditional philology~\cite{xigui1985methods, Lai2019shi, smith2017early, Xing_2011}. To support ACCE, we construct two complementary resources.

First, we construct \textbf{JieZi-Dataset}, the first large-scale, expert-audited VQA training dataset for ACCE. Leveraging an authoritative etymological dictionary~\cite{gu2023hanzi}, we design an expert-in-the-loop pipeline that constrains LLM generation with expert-designed QA templates and dictionary source text, mitigating hallucination while reducing expert effort to template design and stage-wise verification. This pipeline provides over 500K reliable QA pairs and 130K glyph images across multiple script types.

Second, we construct \textbf{JieZi-Bench}, a scholar-grounded evaluation benchmark for ACCE. Its relatively small scale (approximately 8K QA pairs) enables complete construction and verification by human experts, thereby ensuring high evaluation reliability. The reference answers are curated from authoritative lexicographic sources and strictly separated from the training data to prevent leakage. The benchmark is organized into four progressive levels, each aligned with the core dimensions of ACCE.
Experiments on diverse MLLMs show that current models perform well on basic identification but struggle with deeper exegetical tasks such as glyph decomposition, semantic reasoning, and diachronic analysis. Fine-tuning on JieZi-Dataset yields substantial improvements across all four levels, confirming the value of domain-specific training data for this field.
In summary, our contributions are as follows:
\begin{itemize}  [leftmargin=*, nosep]
\item We take the first step toward computational ancient character exegesis (\textbf{ACCE}) by formalizing it as a VQA task with four progressive levels of scholarly analysis.

\item We construct \textbf{JieZi-Dataset}, a collection of 500k expert-audited VQA pairs. We also built the entirely expert-curated \textbf{JieZi-Bench}, providing a standardized evaluation framework for the ACCE task. They were generated by an expert-in-the-loop pipeline that mitigates LLM hallucination via source-grounded constrained generation. 

\item We benchmark representative MLLMs across four progressive levels with explicit reliability metrics, establishing strong baselines and identifying key challenges in deeper scholarly exegesis.
\end{itemize}

\section{Related work}

\subsection{Ancient Chinese Character Datasets}

Recent datasets have assembled large-scale glyph collections for ancient Chinese character recognition, covering tens of thousands of images across oracle bone inscriptions (OBI)~\cite{wang2024open, wang2022oracle}, historical handwritten scripts~\cite{babu2019character, yuan2019large}, and modern printed characters~\cite{zhang2025megahan97k}. These resources have substantially advanced OCR-oriented research. 
However, most of them focus on a single script type with annotations limited to character-class labels. 
Several efforts have begun to enrich annotation dimensions beyond class labels. 
EV-OBC~\cite{guan2024open} introduced a cross-era dataset spanning intermediate scripts such as Small Seal. 
ACCID~\cite{diao2023toward} provided radical-level structural annotations for OBIs, and ACCP~\cite{wang2024puzzle} provided structural and component labels covering characters from multiple eras. OBI Component 20~\cite{hu2024component} provided components of OBIs with expert annotations. 
More recently, PD-OBS~\cite{peng2025interpretable} and OracleSage~\cite{jiang2024oraclesage} took a further step by bridging OBI images with natural-language semantics. 
Despite this progress, such efforts remain confined to oracle bone script, where a significant portion of glyphs remain undeciphered and scholarly consensus on interpretation varies considerably~\cite{li2024comprehensivesurveyoraclecharacter}. 
In contrast, later scripts such as Bronze, Small Seal, and Clerical are supported by substantially more established and reliable scholarly resources.

In summary, existing datasets either provide only sparse symbolic labels or are limited to a single script type with debatable annotations. No dataset offers expert-audited, multi-dimensional natural-language annotations across multiple script types. 
\subsection{Ancient Chinese Character Evaluation}

General vision-language benchmarks such as MMMU~\cite{Yue_2024_CVPR} and DocVQA~\cite{Mathew_2021_WACV} have been instrumental in advancing Multimodal Large Language Models (MLLMs) on document and scene understanding, but they contain no ancient character imagery and therefore cannot assess model capabilities in this domain. Within Chinese cultural heritage, C$^3$-Bench~\cite{cao2024c3benchcomprehensiveclassicalchinese} provides a comprehensive evaluation of classical Chinese cultural knowledge, and MCS-Bench~\cite{liu2025mcs} together with AC-EVAL~\cite{wei-etal-2024-ac} have advanced the assessment of classical text comprehension. However, these benchmarks operate at the passage or knowledge level and do not evaluate single-character visual understanding. In the ancient script domain, OBI-Bench~\cite{chen2024obi} and Oracle-Bench~\cite{qiao2025v} have contributed evaluation frameworks for oracle bone research, including tasks such as fragment matching and visual captioning. However, they focus specifically on oracle bone archaeological scenarios rather than the multi-dimensional exegesis of individual characters across script types. 
Overall, current benchmarks either lack ancient character imagery entirely or assess only recognition accuracy within a single script type, leaving no systematic evaluation for multi-dimensional character exegesis. 

\begin{table*}[ht]
\centering
\scriptsize
\renewcommand{\arraystretch}{1.05}
\caption{Comparison of dataset coverage across ACCE dimensions.
\textbf{MM} denotes whether the dataset provides aligned multimodal evidence beyond plain image-level labels.}
\label{tab:dataset_coverage}
\vspace{-10pt}
\resizebox{\textwidth}{!}{
\begin{tabular}{l cc ccccc c cc c}
\toprule
\multirow{2}{*}{\textbf{Dataset}}
& \multicolumn{2}{c}{\textbf{L1}}
& \multicolumn{5}{c}{\textbf{L2}}
& \multicolumn{1}{c}{\textbf{L3}}
& \multicolumn{2}{c}{\textbf{L4}}
& \multirow{2}{*}{\textbf{MM}} \\
\cmidrule(lr){2-3} \cmidrule(lr){4-8} \cmidrule(lr){9-9} \cmidrule(lr){10-11}
& \textbf{CHAR} & \textbf{SCRC}
& \textbf{STRC} & \textbf{COMR} & \textbf{COMF} & \textbf{COMI} & \textbf{FORC}
& \textbf{ORIM}
& \textbf{COME} & \textbf{EVOI}
& \\
\midrule
OBIMD~\cite{li2026obimd}
& \cmark & \xmark & \xmark & \xmark & \xmark & \xmark & \xmark & \xmark & \xmark & \xmark & \xmark \\

OracleSage~\cite{jiang2024oraclesage}
& \cmark & \xmark & \xmark & \xmark & \xmark & \xmark & \xmark & \xmark & \xmark & \xmark & \cmark \\

ACCP~\cite{wang2024puzzle}
& \cmark & \cmark & \cmark & \cmark & \xmark & \xmark & \xmark & \xmark & \xmark & \xmark & \xmark \\

PD-OBS~\cite{peng2025interpretable}
& \cmark & \cmark & \xmark & \cmark & \xmark & \xmark & \xmark & \cmark & \xmark & \xmark & \cmark \\

\midrule
\textbf{JieZi-Dataset (Ours)}
& \textbf{\cmark} & \textbf{\cmark} & \textbf{\cmark} & \textbf{\cmark} & \textbf{\cmark}
& \textbf{\cmark} & \textbf{\cmark} & \textbf{\cmark} & \textbf{\cmark} & \textbf{\cmark} & \textbf{\cmark} \\
\bottomrule
\end{tabular}
}
\vspace{-10pt}
\end{table*}

\section{Task Definition}

\textbf{Overview.} Ancient Chinese Character Exegesis (ACCE) is a vision-language question answering task. Given an ancient Chinese glyph image and a question $q \in \mathcal{Q}$ about the glyph, the model generates a natural language answer. The question space $\mathcal{Q}$ covers four analytical levels: basic information, glyph form, glyph meaning, and diachronic evolution. ACCE is grounded in established principles of Chinese paleography~\cite{xigui1985methods, Lai2019shi, takashima2021some, Xing_2011} and structures the scholarly exegesis workflow into four progressive levels.

\begin{itemize}[leftmargin=*, nosep]

\item \textbf{L1: Basic Information.} 
This level identifies the core attributes of a glyph through two fundamental tasks. \textbf{Character Recognition (CHAR)} maps the ancient glyph to its modern standard Chinese counterpart. \textbf{Script Classification (SCRC)} identifies the historical script type, such as Oracle Bone, Bronze, Seal, Clerical, or Regular.

\item \textbf{L2: Glyph Form.} 
This level analyzes the internal structure of a glyph. A glyph is decomposed into visual units called components, each with specific structural relationships, functions, and meanings~\cite{bottero1996review}. It encompasses five tasks. \textbf{Structure Classification (STRC)} describes the spatial layout of components. \textbf{Component Recognition (COMR)} identifies the individual components present. \textbf{Component Function (COMF)} explains the functional role of each component. \textbf{Component Interpretation (COMI)} explicates the semantic significance of each component. \textbf{Formation Classification (FORC)} categorizes the character according to the Six Writings taxonomy, including Pictophonetic Characters, Pictographs, and other formation principles.

\item \textbf{L3: Glyph Meaning.} 
This level evaluates the semantic content of a glyph. Ancient Chinese characters carry meaning through abstraction, extension, and historical reinterpretation. It focuses on \textbf{Original Meaning (ORIM)}, which explains the foundational semantic value of a character within its historical context.

\item \textbf{L4: Diachronic Evolution.} 
This level evaluates how a character evolved across historical periods. Characters transform following systematic patterns in graphic simplification, stylistic regularization, and structural reorganization \cite{smith2017early}. It examines the temporal dimension through two tasks. \textbf{Component Evolution (COME)} tracks and explains how specific components changed across script types. \textbf{Evolution Interpretation (EVOI)} provides a holistic analysis of why a character evolved over time, contextualizing these changes within broader patterns of writing system development.
\end{itemize}

Compared to traditional recognition tasks~\cite{zhang2025megahan97k, zhang2020oraclebone,  jiang2023oraclepoints}, ACCE better reflects the goals of paleographic analysis and provides a more realistic testbed for scholar-aligned ancient character understanding.

\section{Datasets}

\begin{figure*}[t]
    \centering
    \includegraphics[width=1\textwidth]{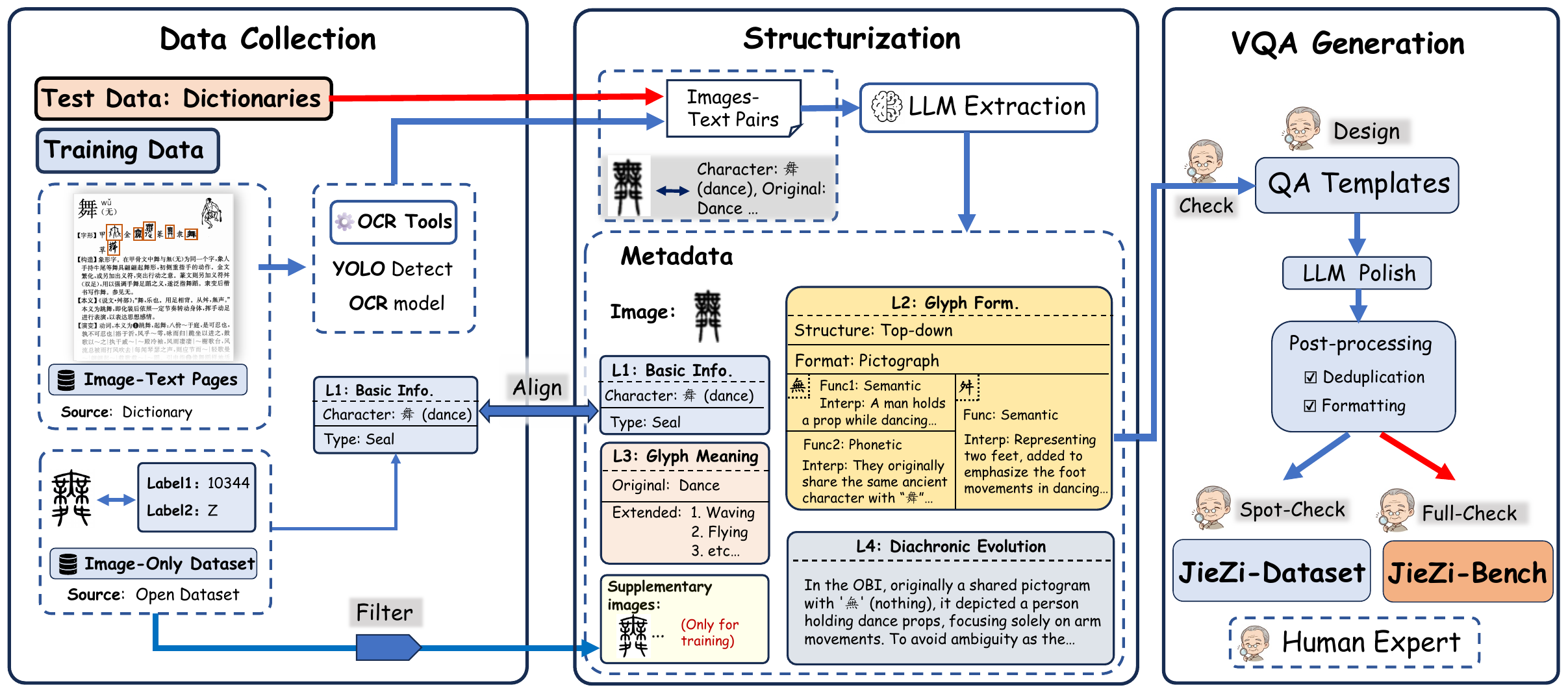} 
    \vspace{-20pt}
    \caption{Data generation pipeline of JieZi-Dataset and JieZi-Bench.}
    \label{fig:pipeline}
\end{figure*}

To support the study of Ancient Chinese Character Exegesis (ACCE), we construct two complementary resources: (1) a large-scale training dataset \textbf{JieZi-Dataset}, and (2) a high-reliability evaluation benchmark \textbf{JieZi-Bench}. Both are built through a shared multi-stage pipeline (Fig.~\ref{fig:pipeline}) but adopt distinct quality control strategies: JieZi-Dataset prioritizes scale through stage-wise spot-checking, while JieZi-Bench ensures evaluation reliability through exhaustive expert verification of every instance.

\subsection{Data Sources}

\textbf{JieZi-Dataset} draws from two sources.
The primary source is a high-resolution scanned edition of \textit{Hanzi Yuanliu Dazidian}~\cite{gu2023hanzi}, a modern etymological dictionary compiled and reviewed by domain experts. The scanned volume exceeds 2K pages and 5M tokens, covering more than 13K characters with rich descriptions of glyph form, meaning, and diachronic evolution across multiple script stages.
To increase glyph diversity, we further incorporate samples from public datasets, including ACCP~\cite{wang2024puzzle} and MegaHan97K~\cite{zhang2025megahan97k}. To ensure annotation consistency, we retain only images whose character identities and script categories align with the corresponding entries.

\textbf{JieZi-Bench} is sourced from four classical and modern lexicographic works: \textit{Kangxi Dictionary}, \textit{Shuowen Jiezi}, \textit{Shuowen Jiezi Zhu}, and \textit{Revised Mandarin Chinese Dictionary}. These dictionaries are selected because their explanations are mutually verifiable and cover characters absent from the training data, preventing data leakage. To ensure sufficient authoritative evidence, we first apply an empirical token-length threshold of 200 to retain the top 20\% most informative entries across all four dictionaries. We then refine the results through expert verification, yielding 1,024 glyph images paired with multi-source explanations.

\subsection{Construction Pipeline}

The construction of both resources follows a three-stage pipeline, as illustrated in Fig.~\ref{fig:pipeline}. The key differences lie in the data collection strategy and the granularity of human verification at each stage.

\textbf{Stage 1: Collection.}
For the \textbf{JieZi-Dataset}, we digitize the scanned dictionary through OCR and glyph extraction. We use a prompt-based OCR pipeline built on Gemini-2.5-Pro~\cite{comanici2025gemini25} to extract textual descriptions from the dictionary. We separately train a YOLOv11~\cite{yolo11_ultralytics} detector on over 2K annotated samples to localize glyph images and classify their script types, covering multiple historical periods and variant forms. Since OCR achieves only ~95\% accuracy, and errors predominantly occur in rare and archaic glyphs that are most critical to ACCE, we manually correct all OCR outputs and extracted glyphs to ensure accuracy, requiring approximately 1,000 hours of human effort. After correction, rule-based normalization produces a coarse-grained alignment between glyph images and their corresponding explanatory text. For external datasets, we use \textit{Hanzi Yuanliu Dazidian} as the reference anchor: ACCP images are retained only when their character and script labels match dictionary entries; MegaHan97K~\cite{zhang2025megahan97k} images are selected for characters covered by the dictionary, restricted to historic document sources to maintain our focus on ancient scripts.
For the \textbf{JieZi-Bench}, data collection centers on cross-dictionary compilation. We merge entries from the four lexicographic sources, apply the token-length threshold described above, and have experts manually verify every retained entry to confirm factual accuracy and cross-source consistency.

\textbf{Stage 2: Structurization.}
Both resources undergo LLM-based structured extraction to convert raw textual entries into standardized metadata records. We employ a prompt-based extraction pipeline with Gemini-3-Flash~\cite{google2025gemini3flashcard} to parse each entry into fields covering \textbf{Basic Information}, \textbf{Glyph Form}, \textbf{Meaning}, and \textbf{Diachronic Evolution}, as illustrated in Fig.~\ref{fig:pipeline}. The extraction prompts for JieZi-Dataset and JieZi-Bench are provided in the supplementary materials.
The verification granularity differs between the two resources. For the JieZi-Dataset, we audit 10\% of the extracted records to ensure the model does not alter, omit, or fabricate content from the original text. For JieZi-Bench, every extracted record is manually checked and revised by experts to guarantee correctness.

\textbf{Stage 3: VQA Generation.}
We design QA templates grounded in real research scenarios from Chinese paleography. For each glyph image, we randomly generate 5 to 10 question-answer pairs, ensuring at least one question from each of the four ACCE levels (L1--L4) to maintain balanced coverage across all subtasks. Post-processing includes removing duplicate images and QA pairs within each task, reformatting lengthy answers into structured markdown for clarity, and verifying image-content alignment.
For JieZi-Dataset, we randomly sample 5K instances for manual inspection. For JieZi-Bench, every QA pair is manually checked and revised to ensure evaluation reliability.

\textbf{Expert-in-the-Loop Quality Assurance.}
Human verification is integrated throughout the pipeline rather than applied as a single final step. At each stage, expert involvement ensures that errors do not propagate downstream. The two resources adopt complementary verification strategies that reflect their distinct roles: JieZi-Dataset employs stage-wise spot-checking to balance scale with quality, while JieZi-Bench applies exhaustive verification at every stage to maximize evaluation reliability.

\begin{figure}[t]
    \centering
    \includegraphics[width=0.9\linewidth]{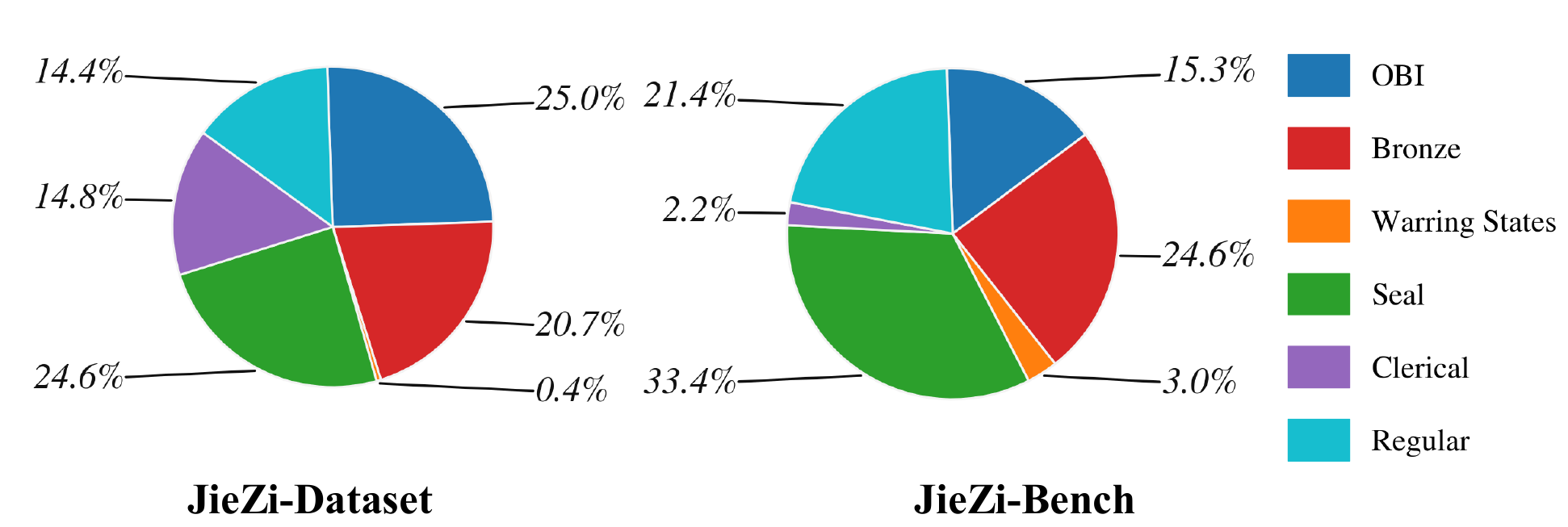}
    \vspace{-10pt}
    \caption{Distribution of glyph stages in JieZi-Dataset and JieZi-Bench.}
    \label{fig:distribution}
    \vspace{-20pt}
\end{figure}

\subsection{Data Statistics}

We report the key statistics of both resources.
JieZi-Dataset comprises approximately 13K unique characters, 130K glyph images spanning six script stages (Oracle Bone, Bronze, Warring States, Seal, Clerical, and Regular), and over 500K expert-audited QA pairs covering all ten ACCE subtasks.
JieZi-Bench, constructed independently from separate lexicographic sources, contains 1,024 glyph images and approximately 8K QA pairs, with every instance verified by human experts.

\textbf{Task coverage.}
A notable advantage of JieZi-Dataset is its complete coverage of all ACCE dimensions.
As shown in Tab.~\ref{tab:dataset_coverage}, existing datasets address at most four of the ten subtasks, and only OracleSage~\cite{jiang2024oraclesage} and PD-OBS~\cite{peng2025interpretable} provide aligned multimodal evidence beyond image-level labels.
In contrast, JieZi-Dataset is the first resource to support the full exegesis workflow, encompassing basic identification (L1), glyph-form analysis (L2), meaning exegesis (L3), and diachronic evolution (L4) within a unified multimodal framework.

\begin{figure}[t]
    \centering
    \includegraphics[width=\linewidth]{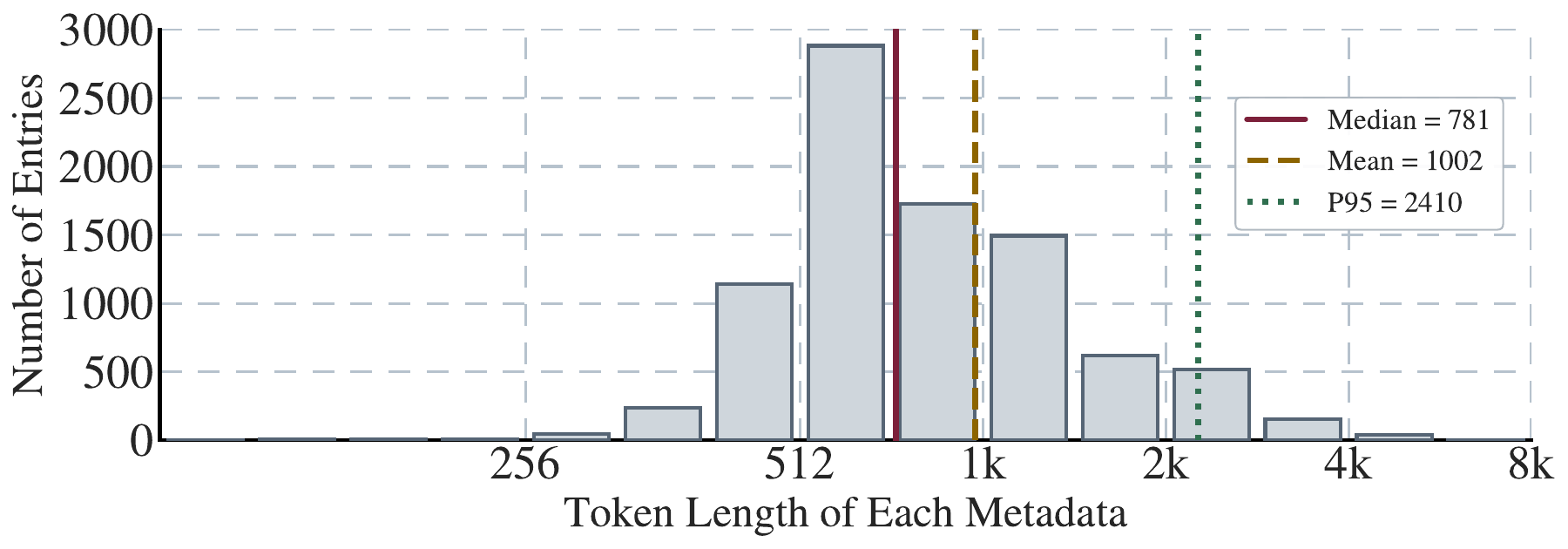}
    \vspace{-20pt}
    \caption{Distribution of token lengths across metadata entries
in the dataset. }
    \label{fig:tokendistribution}
    \vspace{-15pt}
\end{figure}

\textbf{Annotation richness.}
In addition to broad task coverage, JieZi-Dataset provides substantially richer per-character annotations than prior resources.
As shown in Fig.~\ref{fig:tokendistribution}, the median per-entry metadata length is 781 tokens and the mean reaches 1,002 tokens, with the majority of entries falling in the 512 to 1,024 token range.
In contrast, existing datasets typically rely on single categorical labels or short phrases. Each character in JieZi-Dataset, however, is accompanied by detailed, multi-dimensional textual descriptions, making it suitable for training generative models.

\begin{figure}[t]
    \centering
    \includegraphics[width=\linewidth]{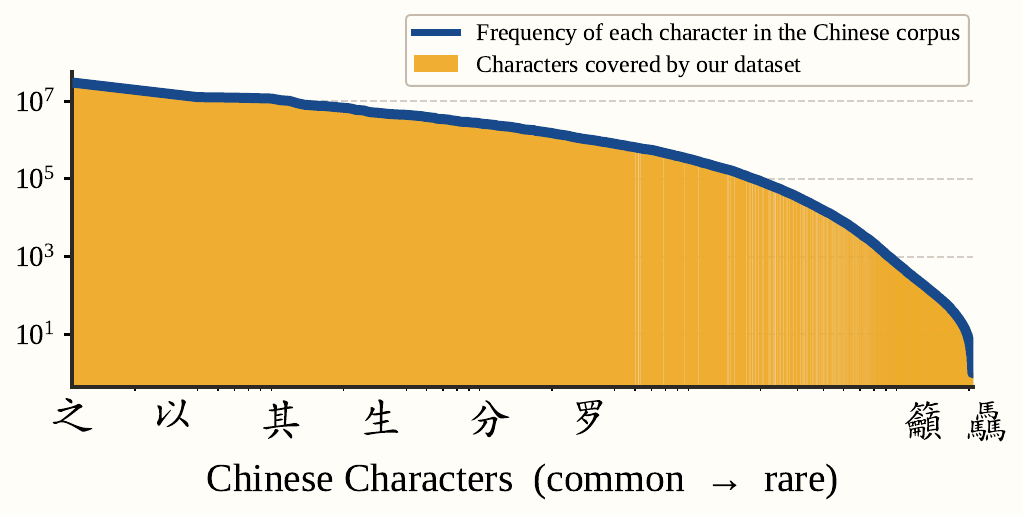}
    \vspace{-20pt}
    \caption{Coverage of JieZi-Dataset over character frequencies in a classical Chinese corpus. 
    }
    \label{fig:frequency}
    \vspace{-15pt} 
\end{figure}

\textbf{Character and script diversity.}
We further assess whether the dataset adequately represents the diversity of real-world ancient texts.
Fig.~\ref{fig:frequency} plots the character-frequency distribution of a representative classical Chinese corpus~\cite{shim5hisdoc}.
Although this corpus exhibits a pronounced long-tail pattern, JieZi-Dataset covers a substantial portion of both high-frequency and lower-frequency characters, thereby ensuring broad applicability to downstream tasks.
Fig.~\ref{fig:distribution} further presents the glyph-stage composition of JieZi-Dataset and JieZi-Bench.
Both resources span all six script stages rather than concentrating on a single type, with Seal and Bronze scripts constituting a significant proportion.
Notably, 12.3\% of glyph images are sourced from real historical documents with natural degradation such as erosion and stains, rather than clean dictionary renderings.
In the construction of JieZi-Bench, we introduced some splits to better reflect the generalization ability of JieZi-Dataset. First, 25\% of the images in JieZi-Bench belong to unseen (character, script) pairs that do not appear in JieZi-Dataset. Second, 12.5\% of the characters in JieZi-Bench are entirely unseen during training. Third, 25.2\% of the components in JieZi-Bench are unseen in JieZi-Dataset.

\section{Experiment}

\subsection{Experimental Setup}

To comprehensively evaluate the ACCE task, we benchmark existing SOTA MLLMs, including closed-source commercial models such as GPT-5.4~\cite{gpt54thinking_system_card} and open-source models such as Qwen3.5-397b-a17b~\cite{qwen3.5}.
In addition, we validate the effectiveness of our high-quality training data on Qwen3.5-2B, Qwen3.5-4B and Qwen3.5-9B.
The training is performed on 8 Ascend 910B NPUs.
More training details are presented in the supplementary material.

\subsection{Evaluation Metrics} 
We use different metrics for closed-form and open-form questions.

\begin{itemize}[leftmargin=*]
    \item \textbf{Acc (Accuracy):} For tasks with categorical outputs (CHAR, SCRC, STRC, FORC), we use accuracy.
    \item \textbf{F1-Score:} For component identification and functional description tasks (COMR, COMF, COME), we use character-level F1-score~\cite{vanrijsbergen1979information}, which better reflects real-world exegesis scenarios, where each glyph is treated as a single instance and component predictions may be only partially correct.
    \item \textbf{BERTScore:} For open-ended natural-language generation tasks (ORIM, COMI), we use BERTScore~\cite{zhangbertscore} to measure semantic similarity between model outputs and references.
    \item \textbf{LLM-as-a-Judge:} For EVOI, an open-ended generation task, we adopt an LLM-as-a-Judge protocol. The judge evaluates responses along two dimensions: (1) \textbf{Fact Alignment}, measuring consistency with the reference answer; (2) \textbf{Scholarly Expression}, assessing the use of appropriate domain-specific terminology. Full prompts and the human validation study are provided in the supplementary material.
\end{itemize}

\begin{table*}[t]
\centering
\small
 \setlength{\tabcolsep}{4.2pt}
 \renewcommand{\arraystretch}{1.12}
\caption{Main results on JieZi-Bench across four paleographic levels: \textbf{L1} (Basic Info: CHAR, SCRC), \textbf{L2} (Glyph Form: STRC, COMR, COMF, COMI, FORC), \textbf{L3} (Meaning: ORIM), and \textbf{L4} (Evolution: COME, EVOI-FAC/SCE). All metrics are scaled to 0-100.}
\label{tab:main_results}
\resizebox{\textwidth}{!}{
\begin{tabular}{l cccccccc c cc}
\toprule
\multirow{3}{*}{\textbf{Method}}
& \multicolumn{2}{c}{\textbf{L1}}
& \multicolumn{5}{c}{\textbf{L2}}
& \multicolumn{1}{c}{\textbf{L3}}
& \multicolumn{3}{c}{\textbf{L4}} \\
\cmidrule(lr){2-3} \cmidrule(lr){4-8} \cmidrule(lr){9-9} \cmidrule(lr){10-12}
& \multirow{2}{*}{\textbf{CHAR}$\uparrow$} & \multirow{2}{*}{\textbf{SCRC}$\uparrow$}
& \multirow{2}{*}{\textbf{STRC}$\uparrow$} & \multirow{2}{*}{\textbf{COMR}$\uparrow$} & \multirow{2}{*}{\textbf{COMF}$\uparrow$} & \multirow{2}{*}{\textbf{COMI}$\uparrow$} & \multirow{2}{*}{\textbf{FORC}$\uparrow$}
& \multirow{2}{*}{\textbf{ORIM}$\uparrow$}
& \multirow{2}{*}{\textbf{COME}$\uparrow$}
& \multicolumn{2}{c}{\textbf{EVOI}} \\
\cmidrule(lr){11-12}
& & & & & & & & & & \textbf{FAC}$\uparrow$ & \textbf{SCE}$\uparrow$ \\
\midrule
\multicolumn{12}{l}{\textit{Closed-source MLLMs}} \\
GPT-5.4 Thinking~\cite{gpt54thinking_system_card}
& 10.4 & 39.2
& 50.4 & 13.8 & 11.9 & 10.3 & 33.1
& 59.9
& 12.4 & 10.2 & 16.8 \\
Gemini 3.1 Pro~\cite{gemini31pro_model_card}
& 29.9 & 76.6
& 65.7 & 31.9 & 29.9 & 24.8 & 56.1
& 62.8
& 30.3 & 22.3 & 27.1 \\
Claude opus 4.6~\cite{claude_opus46_system_card}
& 22.9 & 65.8
& 63.5 & 28.0 & 26.3 & 22.3 & 50.6
& 59.3
& 26.7 & 19.8 & 26.3 \\
Doubao-Seed-2.0-pro~\cite{seed20_model_card}
& 36.6 & 68.5
& 69.8 & 37.8 & 34.4 & 29.9 & 59.9
& 63.2
& 36.1 & 31.7 & 38.7 \\
\midrule
\multicolumn{12}{l}{\textit{Open-source MLLMs}} \\
Kimi-K2.5~\cite{kimiteam2026kimik25visualagentic}
& 31.5 & 70.1
& 68.8 & 39.2 & 36.6 & 31.5 & 58.1
& 66.6
& 36.4 & 28.3 & 34.6 \\
GLM-4.6V~\cite{glm46v_paper}
& 18.2 & 49.7
& 48.9 & 21.8 & 20.4 & 18.2 & 42.2
& 48.2
& 20.9 & 15.5 & 20.2 \\
Qwen3.5-397b-a17b~\cite{qwen3.5}
& 26.2 & 67.1
& 69.4 & 32.1 & 29.9 & 26.5 & 54.4
& 59.5
& 28.3 & 22.2 & 30.7 \\
\midrule
Qwen3.5-2B~\cite{qwen3.5}
& 18.9 & 22.6
& 31.6 & 17.3 & 13.7 & 17.0 & 30.7
& 62.3
& 12.9 & 7.9 & 20.2 \\
\textbf{Qwen3.5-2B + JieZi-Dataset}
& 41.8\up{22.9} & 61.7\up{39.1}
& 72.7\up{41.1} & 42.6\up{25.3} & 32.7\up{19.0} & 41.3\up{24.3} & 63.3\up{32.6}
& 65.7\up{3.4}
& 28.3\up{15.4} & 27.9\up{20.0} & 36.7\up{16.5} \\
\midrule
Qwen3.5-4B~\cite{qwen3.5}
& 3.6 & 6.6
& 1.0 & 4.5 & 3.9 & 4.3 & 5.9
& 6.5
& 3.6 & 2.0 & 3.6 \\
\textbf{Qwen3.5-4B + JieZi-Dataset}
& 49.0\up{45.4} & 67.6\up{61.0}
& 74.1\up{73.1} & 47.2\up{42.7} & 37.9\up{34.0} & 42.4\up{38.1} & 64.8\up{58.9}
& 65.9\up{59.4}
& 33.2\up{29.6} & 28.5\up{26.5} & 37.2\up{33.6} \\
\midrule
Qwen3.5-9B~\cite{qwen3.5}
& 22.3 & 51.0
& 40.2 & 27.4 & 24.6 & 26.1 & 49.1
& 64.7
& 22.4 & 15.0 & 23.5 \\
\textbf{Qwen3.5-9B + JieZi-Dataset}
& \textbf{52.6}\up{30.3} & \textbf{80.1}\up{29.1}
& \textbf{74.7}\up{34.5} & \textbf{48.1}\up{20.7} & \textbf{38.4}\up{13.8} & \textbf{45.7}\up{19.6} & \textbf{66.5}\up{17.4}
& \textbf{66.7}\up{2.0}
& \textbf{45.4}\up{23.0} & \textbf{32.5}\up{17.5} & \textbf{40.6}\up{17.1} \\
\bottomrule
\end{tabular}
}
\end{table*}

\subsection{Main Results}
We benchmark several general MLLMs against models fine-tuned on JieZi-Dataset to evaluate the necessity of domain-specific data. The ablation study of our proposed pipeline and dataset is present in the supplementary material. Tab.~\ref{tab:main_results} reveals the following insights:

\textbf{General MLLMs show varying performance across all ACCE levels.}
Most models perform moderately on categorical tasks (e.g., SCRC 40--77\%) but struggle with fine-grained analysis (e.g., CHAR 10--37\%). Performance deteriorates further on L4, where even the best non-fine-tuned model scores below 40\% on EVOI. This indicates that general pretraining fails to encode the structured paleographic knowledge required for exegesis.

\textbf{Models trained on more Chinese data perform notably better.}
Among non-fine-tuned models, Doubao-Seed-2.0-pro and Kimi-K2.5 rank as the top two across nearly all subtasks. Both models are developed by Chinese technology companies and are likely trained on richer Chinese and classical-text data. In contrast, GPT-5.4 lags far behind despite its strong general capabilities (CHAR: 10.4\% vs.\ Doubao's 36.6\%). This suggests that domain-relevant data coverage, rather than model scale, is a primary bottleneck for ACCE.

\textbf{Domain-specific fine-tuning consistently improves performance, and the improvement increases with model capacity.}
Fine-tuning on JieZi-Dataset improves performance across all subtasks. Even the lightweight 2B model outperforms Gemini-3.1-Pro and GPT-5.4 on structural parsing (e.g., COMR and FORC), and the 9B model achieves SOTA results across the board. Performance further improves from 2B to 9B, with larger gains on deeper reasoning tasks (e.g., COME: +4.9 from 2B to 4B, +12.2 from 4B to 9B), suggesting that further scaling remains a promising direction. 

\subsection{Generalization Analysis}

We analyze the robustness and generalization of the fine-tuned model on JieZi-Bench, which includes \textbf{Unseen Characters~(UC)} and \textbf{Unseen Glyphs~(UG)}. Tab.~\ref{tab:generalization_by_glyph} reports results for Qwen3.5-9B fine-tuned on JieZi-Dataset across Bronze, Seal, and Regular scripts.

\textbf{Exegesis remains robust despite recognition failures.}
CHAR falls to near-zero on unseen subsets (e.g., Seal UC: 1.9\%, Bronze UG: 5.7\%), indicating that exact character identification does not generalize to novel glyphs. However, structural parsing metrics show no comparable collapse: from All to UG, COMF decreases by only 3.5 points on Bronze (14.6$\to$11.1), 2.1 on Seal (51.9$\to$49.8), and 7.8 on Regular (69.6$\to$61.8). This decoupling confirms that training on JieZi-Dataset induces transferable paleographic knowledge: the model derives structural and semantic understanding from visual form rather than relying on overfitting to seen character identities.

\textbf{Older scripts remain the most challenging.}
Across nearly all metrics, performance decreases from Regular to Seal to Bronze (e.g., CHAR: 82.2$\to$51.5$\to$18.3; COMR: 85.1$\to$65.5$\to$17.4). The greater visual variance and structural abstraction of earlier scripts pose a persistent challenge for future research.

\begin{table}[t]
  \centering
  \scriptsize
  \renewcommand{\arraystretch}{1.12}
  \caption{Generalization results (\%) across Bronze, Seal, and Regular scripts. \textbf{UC}: Unseen Characters. \textbf{UG}: Unseen Glyphs. \textbf{All}: Full split. Metrics are scaled to 0-100.}
  \label{tab:generalization_by_glyph}
  \resizebox{\columnwidth}{!}{
  \begin{tabular}{l ccc ccc ccc}
    \toprule
    \textbf{Metric} & \multicolumn{3}{c}{\textbf{Bronze}} & \multicolumn{3}{c}{\textbf{Seal}} & \multicolumn{3}{c}{\textbf{Regular}} \\
    \cmidrule(lr){2-4} \cmidrule(lr){5-7} \cmidrule(lr){8-10}
     & \shortstack{UC\\$n$=26} & \shortstack{UG\\$n$=87} & \shortstack{All\\$n$=252} & \shortstack{UC\\$n$=53} & \shortstack{UG\\$n$=56} & \shortstack{All\\$n$=342} & \shortstack{UC\\$n$=31} & \shortstack{UG\\$n$=53} & \shortstack{All\\$n$=219} \\
    \midrule
    CHAR & 7.7 & 5.7 & 18.3 & 1.9 & 1.8 & 51.5 & 6.5 & 37.7 & 82.2 \\
    SCRC & 56.2 & 54.8 & 51.7 & 90.6 & 87.5 & 88.3 & 93.5 & 94.3 & 98.6 \\
    \midrule
    FORC & 48.1 & 42.9 & 48.5 & 82.5 & 81.7 & 78.4 & 87.9 & 90.1 & 87.7 \\
    STRC & 57.7 & 54.0 & 58.7 & 86.8 & 85.7 & 81.6 & 93.5 & 92.5 & 93.2 \\
    COMR & 20.1 & 12.8 & 17.4 & 64.0 & 63.2 & 65.5 & 60.8 & 71.2 & 85.1 \\
    COMF & 16.3 & 11.1 & 14.6 & 50.8 & 49.8 & 51.9 & 57.5 & 61.8 & 69.6 \\
    COMI & 20.1 & 12.8 & 16.9 & 64.0 & 63.2 & 64.7 & 49.5 & 56.1 & 76.7 \\
    \midrule
    ORIM & 61.8 & 60.1 & 61.1 & 68.1 & 67.6 & 68.7 & 74.5 & 74.8 & 75.2 \\
    \midrule
    COME & 15.5 & 9.5 & 12.8 & 46.6 & 46.0 & 48.3 & 45.6 & 53.3 & 63.8 \\
    FAC & 10.6 & 6.9 & 12.7 & 33.0 & 32.6 & 41.2 & 42.7 & 46.2 & 54.2 \\
    SCE & 19.2 & 14.7 & 22.6 & 47.2 & 46.4 & 49.5 & 64.5 & 61.3 & 67.4 \\
    \bottomrule
  \end{tabular}
  }
\end{table}

\section{Conclusion}

In this work, we introduce Ancient Chinese Character Exegesis (ACCE), a vision-language task that structures paleographic analysis into four progressive levels: basic information, glyph form, meaning, and diachronic evolution.
To support this task, we construct JieZi-Dataset, a large-scale expert-audited dataset with approximately 500K QA pairs derived from authoritative etymological sources, and JieZi-Bench, a scholar-grounded evaluation benchmark aligned with the same four-level structure. 
Experiments show that current MLLMs perform reasonably well on basic identification but struggle with deeper exegetical tasks such as glyph decomposition and diachronic reasoning.
Fine-tuning on JieZi-Dataset yields substantial improvements across all four levels, confirming the critical role of domain-specific data.
Our work contributes the first resource covering the complete exegesis workflow across multiple script types, establishing a standardized foundation and a reproducible baseline for computational paleography.
Building on this foundation, this effort facilitates broader exploration at the intersection of artificial intelligence and ancient Chinese character studies, paving the way for more robust, interpretable, and domain-aligned analytical tools.

\bibliographystyle{ACM-Reference-Format}

\bibliography{references.bib}

\clearpage
\appendix
\section*{Supplementary Material}

This supplementary material provides additional details that could not be included in the main paper due to space constraints. First, Sec.~\ref{sec:task_details} provides scholarly background on the ACCE task formulation and illustrative examples.
Second, Sec.~\ref{sec:data_details} presents further details on dataset construction, including data distribution, prompt templates, and quantitative verification statistics. Finally, Sec.~\ref{sec:exp_details} provides additional experimental details and the complete results on JieZi-Bench.

\section{Task Formulation Details}
\label{sec:task_details}

\subsection{Scholarly Background of ACCE}
\label{sec:background}

\textbf{Ancient character exegesis.}
In Chinese paleography, \textit{exegesis} refers to the scholarly practice of interpreting an ancient glyph by jointly analyzing its visual form, internal structure, semantic content, and historical evolution~\cite{xigui1985methods, Lai2019shi, Xing_2011}. As one of the oldest continuously used writing systems, ancient Chinese characters carry irreplaceable value for research in archaeology, history, and historical linguistics~\cite{denecke2017oxford, keightley1996art}, making their systematic interpretation a long-standing scholarly priority. Unlike modern character recognition, which maps a glyph to a single Unicode label, exegesis requires the scholar to explain \textit{why} a character has its particular form, meaning, and evolutionary trajectory. In practice, scholarly exegesis spans four complementary analytical dimensions~\cite{bottero1996review, smith2017early}, which are examined jointly and iteratively rather than in a fixed sequence:
\begin{itemize}[leftmargin=*, nosep]
\item \textbf{Identification.} \textbf{Identifying the character} and \textbf{determining which historical script period} it belongs to.
\item \textbf{Structural Analysis.} Analyzing the glyph's \textbf{internal structure} by \textbf{decomposing it into components} and examining how each component contributes to the character's form and meaning.
\item \textbf{Semantic Interpretation.} Interpreting the character's \textbf{original semantic value} in its historical context, informed by the structural analysis.
\item \textbf{Diachronic Tracing.} Tracing how both \textbf{form and meaning evolved} across successive script periods, and explaining the motivations behind these diachronic changes.
\end{itemize}
This multi-faceted, knowledge-intensive analysis constitutes the standard methodology taught in Chinese philology programs and practiced in archaeological and linguistic research.

\begin{figure}
    \centering
    \includegraphics[width=1\linewidth]{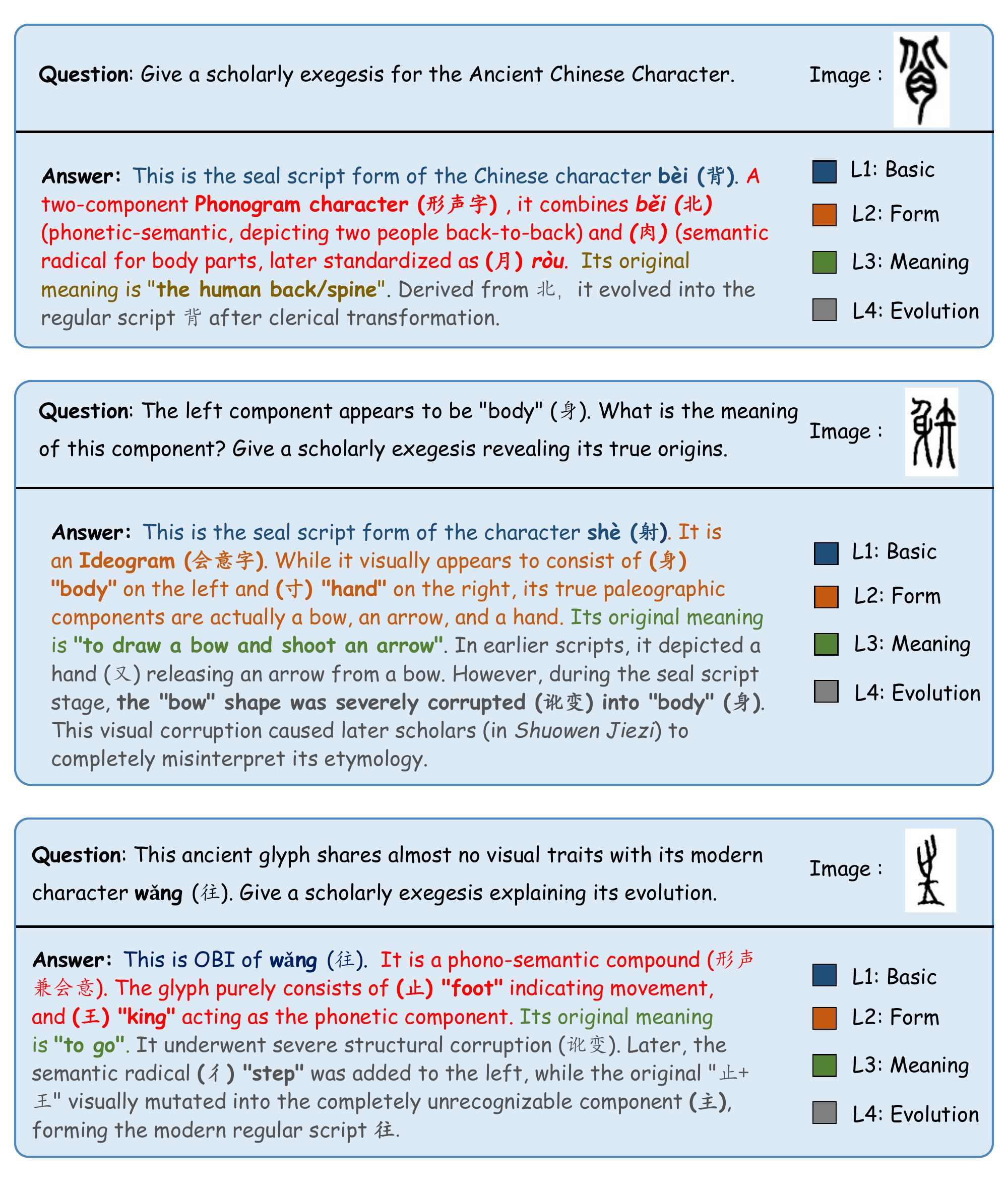}
    \caption{Examples of comprehensive scholarly exegesis in complex paleographic scenarios.}
    \label{fig:full_task_example}
\end{figure}

\textbf{Challenges in exegesis.}
Exegesis is particularly challenging because it demands the simultaneous integration of heterogeneous knowledge types. These include visual pattern recognition for identifying glyph components, linguistic knowledge for understanding component functions, semantic reasoning for inferring original meaning, and historical knowledge for explaining diachronic change. Each requires different cognitive skills and a distinct body of expertise.
Moreover, ancient glyphs exhibit high visual variability. The same character may appear markedly different across script periods (e.g., OBI, Bronze, Seal, Clerical), and different characters may share visually similar components, making identification and structural analysis error-prone even for trained scholars.
The deeper analytical dimensions are also interrelated: a structural misjudgment (e.g., misidentifying a component) propagates into incorrect semantic and evolutionary interpretations. This interdependence among form, meaning, and evolution is why holistic exegesis cannot be reduced to a single classification step, although each dimension can still be evaluated independently. Training a human expert in character exegesis typically requires years of graduate-level study, and the analysis of a single character may involve cross-referencing multiple authoritative dictionaries and reconciling conflicting scholarly interpretations~\cite{chapinal2023review, shen2020phonological}.

\begin{figure}
    \centering
    \includegraphics[width=1\linewidth]{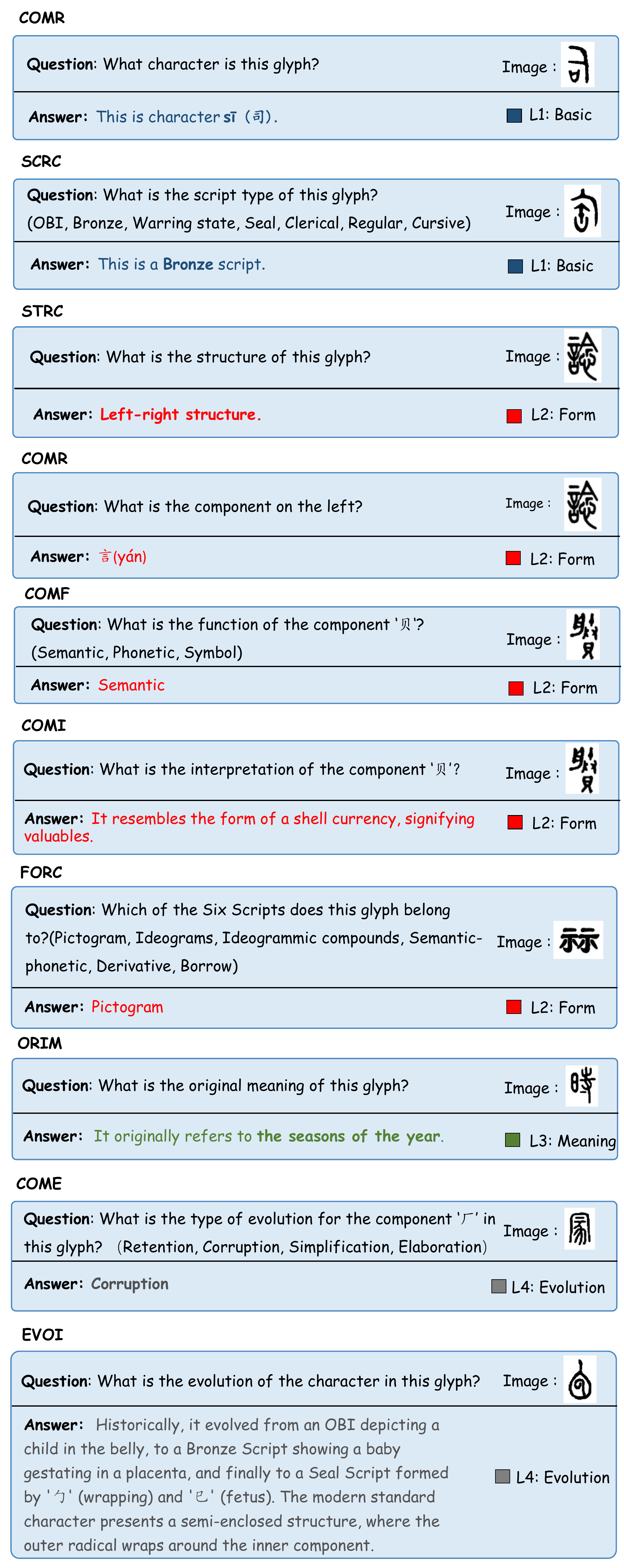}
    \caption{Examples of Question-Answering pairs across the ten fine-grained sub-tasks within the ACCE framework.}
    \label{fig:subtask}
\end{figure}

\textbf{Formalization as a computational task.}
Despite its importance, exegesis has received limited attention from the computational community. Existing work on ancient Chinese characters focuses predominantly on recognition and retrieval~\cite{chi-etal-2024-ancient, li2024comprehensivesurveyoraclecharacter}, which corresponds to only the most elementary step of the full exegetical workflow. To our knowledge, no prior work has attempted to formalize the complete, multi-stage scholarly process as a computational task.
ACCE addresses this gap by decomposing exegesis into four analytical levels that mirror these dimensions: identifying a character and its script period (\textbf{L1: Basic Information}), analyzing internal structure and component functions (\textbf{L2: Glyph Form}), interpreting the original semantic value (\textbf{L3: Glyph Meaning}), and tracing diachronic evolution (\textbf{L4: Diachronic Evolution}). The levels are ordered from basic to deeper analysis for clarity, but they are not a strict pipeline: each is independently answerable from the glyph image, while the deeper levels (L2--L4) remain conceptually interrelated. It enables researchers to construct targeted supervision, apply stage-specific evaluation metrics, and identify precisely which aspects of exegesis current models can and cannot handle.

\begin{figure}[t]
    \centering
    \includegraphics[width=1\linewidth]{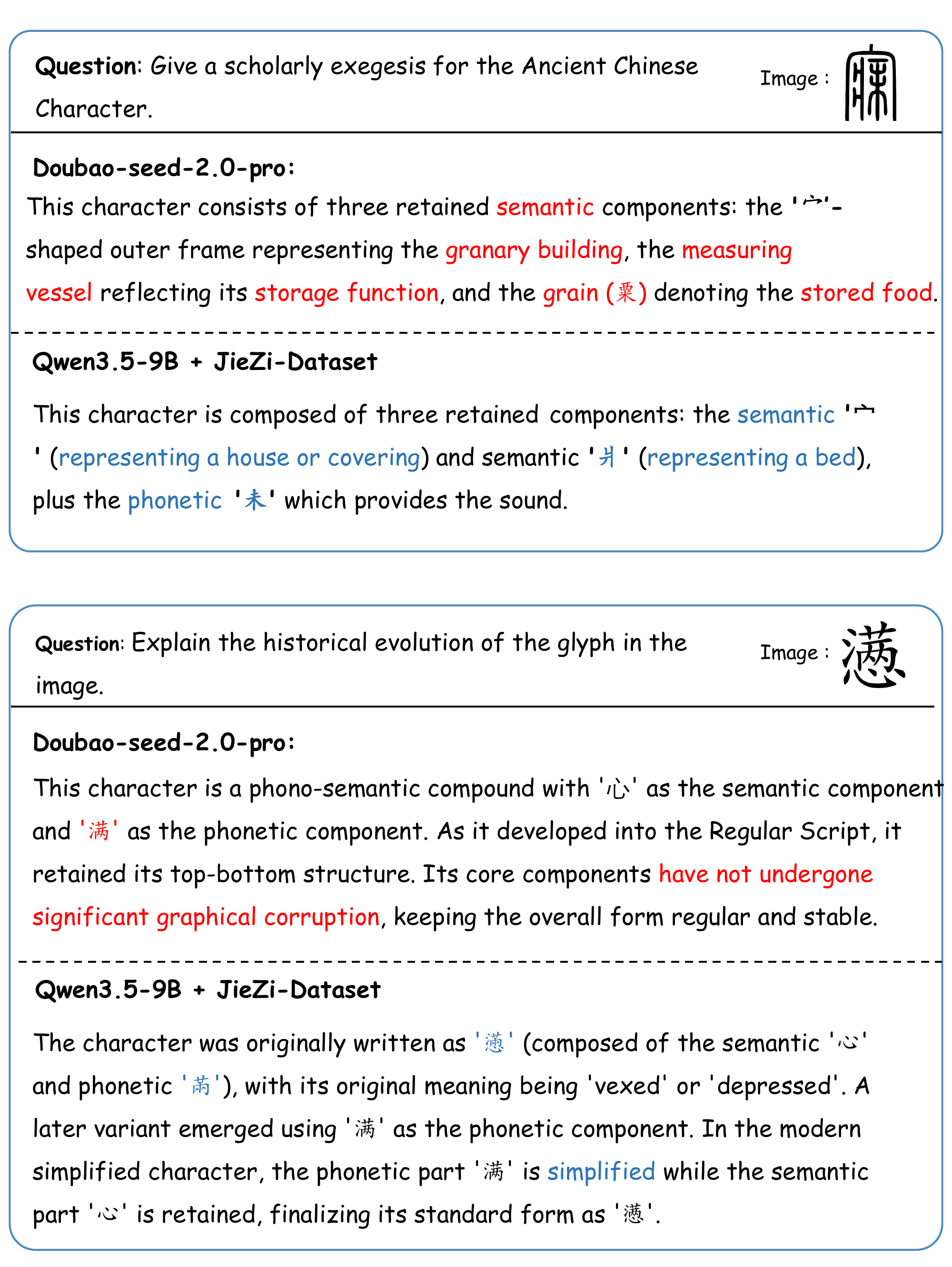}
    \caption{Examples of responses for different tasks from different models. Red text indicates error messages, and blue text indicates correct messages.}
    \label{fig:contrast}
\end{figure}

\subsection{Examples of Tasks}
\label{sec:task_examples}

To provide a more intuitive understanding of the ACCE task, we present qualitative examples from the JieZi-Dataset. Fig.~\ref{fig:subtask} illustrates the ten fine-grained subtasks. While these subtasks assess individual capabilities, the ultimate goal of ACCE is holistic interpretation. Fig.~\ref{fig:full_task_example} presents comprehensive scholarly exegesis in real scenarios, where models are challenged with complex paleographic phenomena such as severe structural corruption and visually similar but semantically distinct glyph forms. As demonstrated, a high-quality exegesis successfully resolves these challenges by synthesizing all four progressive levels (L1 to L4) into a coherent, domain-aligned explanation.
Fig.~\ref{fig:contrast} contrasts the responses of the fine-tuned model with those of baseline models on representative subtasks.

\section{Dataset Construction and Quality Control}
\label{sec:data_details}

\subsection{Data Distribution and Metrics}
\label{sec:metric_details}

\textbf{JieZi-Dataset}.
To illustrate the data diversity and comprehensiveness, Tab.~\ref{tab:dataset_coverage_supp} presents the subtask coverage distribution within the JieZi-Dataset. Unlike the individually targeted test queries in JieZi-Bench, the training data often features comprehensive QAs that simultaneously address multiple analytical dimensions, mirroring the holistic nature of real-world scholarly exegesis. Consequently, the sum of subtask frequencies exceeds the total number of unique QA pairs.

\begin{table}[t]
\centering
\caption{Coverage of ACCE subtasks within the JieZi-Dataset. Since a single comprehensive QA pair may simultaneously address multiple subtasks, the cumulative count across all subtasks exceeds the total number of unique QA pairs (500K).}
\label{tab:dataset_coverage_supp}
\begin{tabular}{llc}
\toprule
\textbf{Level} & \textbf{Subtask} & \textbf{QAs Covering Subtask} \\
\midrule
\multirow{2}{*}{L1: Basic} & CHAR & 175,875\\
 & SCRC & 175,875\\
\midrule
\multirow{5}{*}{L2: Form} & STRC & 175,875\\
 & COMR & 175,875\\
 & COMF & 221,558\\
 & COMI & 221,558\\
 & FORC & 312,984\\
\midrule
L3: Meaning & ORIM & 175,875\\
\midrule
\multirow{2}{*}{L4: Evolution} & COME & 221,558\\
 & EVOI & 210,360\\
\midrule
\textbf{Overall} & \textbf{Unique QA Pairs} &   505,867 \\
\bottomrule
\end{tabular}
\end{table}

\textbf{JieZi-Bench}.
Tab.~\ref{tab:question_distribution} summarizes the question distribution and corresponding evaluation metrics across the ACCE tasks in JieZi-Bench. Notably, the four component-related subtasks (COMR, COMF, COMI, and COME) share the same set of test questions (1,852 per subtask), because each question requires the model to analyze all four component dimensions simultaneously. Within this shared question set, each subtask is evaluated with a distinct metric.

\begin{table}
\centering
\caption{Distribution of Each Task in JieZi-Bench}
\label{tab:question_distribution}
\begin{tabular}{llc}
\toprule
\textbf{Task} & \textbf{Metric} & \textbf{Question Count} \\
\midrule
CHAR & Accuracy & 1024 \\
SCRC & Accuracy & 1024 \\
STRC & Accuracy & 1024 \\
FORC & Accuracy & 1024 \\
ORIM & BERTScore & 1024 \\
EVOI & LLM-as-a-Judge & 1024 \\
COMR & F1-Score + Accuracy & 1852 \\
COMF & F1-Score + Accuracy & 1852 \\
COMI & F1-Score + BERTScore & 1852 \\
COME & F1-Score + Accuracy & 1852 \\
\midrule
\textbf{Overall} & \textbf{Unique QA Pairs} & 7996 \\
\bottomrule
\end{tabular}
\end{table}

\subsection{Prompt for Structurization}
\label{sec:data_prompts}

Raw dictionary entries are written in dense, unstructured prose that interleaves character identity, structural analysis, semantic explanation, and evolutionary commentary in a single paragraph. Stage~2 (Structurization) aims to distill this unstructured text into compact, structured JSON records aligned with the four ACCE levels (L1--L4), retaining only the information relevant to exegesis while discarding editorial remarks, cross-references, and other content not directly useful for downstream VQA generation.

This structurization is non-trivial because ancient Chinese characters exhibit highly heterogeneous complexity. A simple pictograph may require only a brief structural note, whereas a phonosemantic compound spanning multiple script periods may involve dozens of components, variant forms, and evolutionary branches. The prompt must therefore accommodate this wide range of complexity within a single unified schema without losing critical details for complex entries or generating spurious fields for simple ones.

\begin{figure*}
    \centering
    \includegraphics[width=0.94\textwidth]{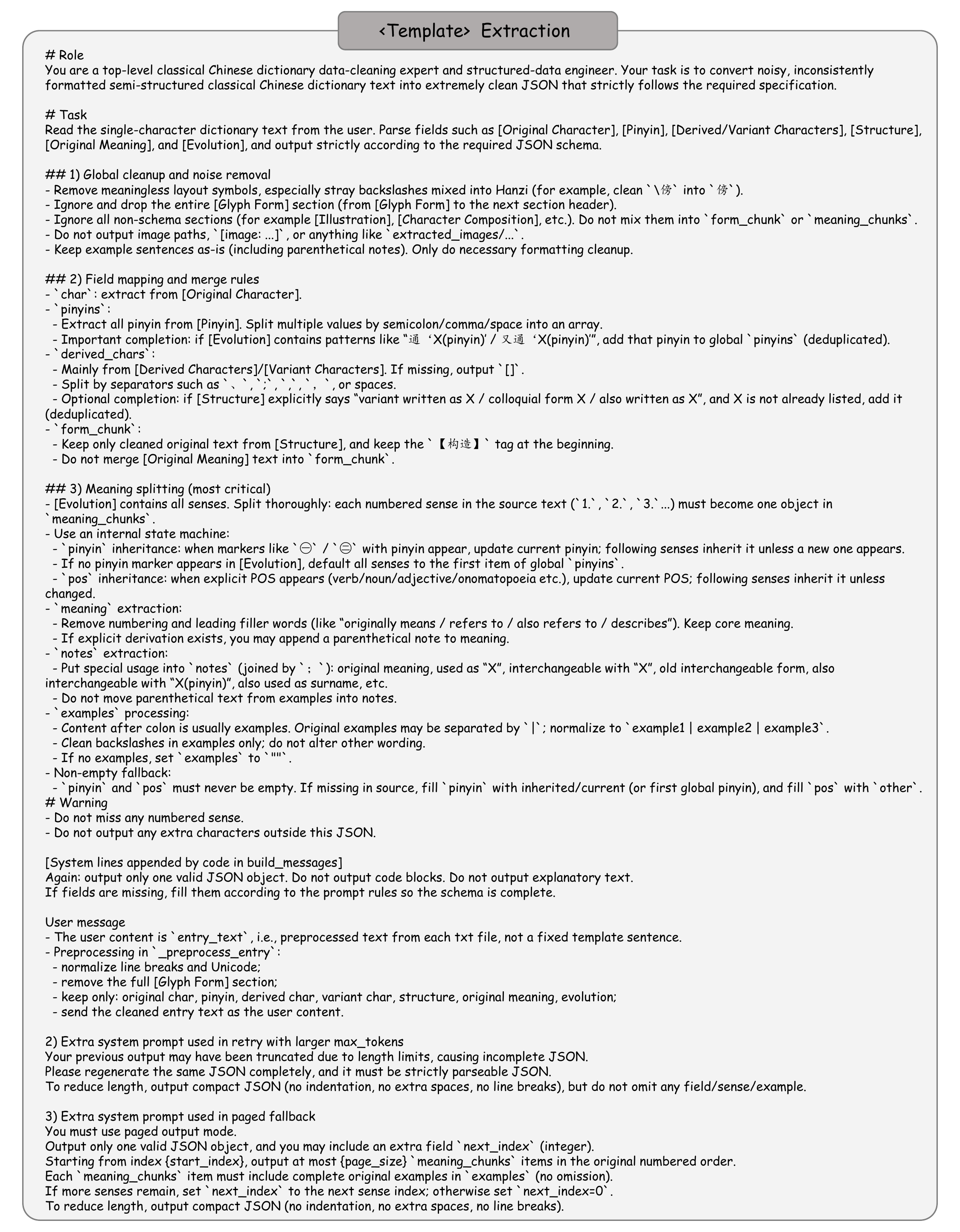}
    \caption{Prompt template used for structured metadata extraction from dictionary entries.}
    \label{fig:extract-prompt}
\end{figure*}

Fig.~\ref{fig:extract-prompt} presents the full prompt used in this stage. Its design addresses three key requirements:
(1)~\textit{Schema formatting}. The prompt specifies a strict JSON schema with all required field names and value types, ensuring machine-parseable outputs without post-hoc reformatting.
(2)~\textit{Source fidelity}. The model is explicitly instructed to extract only from the provided text and to leave fields empty rather than fabricate content, which is critical for mitigating hallucination on rare or ambiguous entries.
(3)~\textit{Adaptive granularity}. The schema naturally accommodates entries of varying complexity: simple characters produce compact records, while complex multi-period entries expand as needed without requiring separate templates.

\subsection{Templates for VQA generation}

Stage~3 (VQA Generation) transforms the structured metadata from Stage~2 into natural-language VQA pairs suitable for supervised fine-tuning. The generated QA pairs are designed to \textit{simulate the progressive reasoning process of domain experts}, reflecting the layered analytical workflow of scholarly exegesis rather than isolated factual retrieval.

We design four prompt templates corresponding to the four ACCE levels (L1--L4) and adopt a \textit{dual-track generation strategy}.
For \textbf{factual subtasks} with well-defined answer patterns (e.g., character identity at L1, structure type at L2), QA pairs are produced by directly instantiating the template with the corresponding metadata fields, requiring no LLM reasoning.
For \textbf{reasoning-intensive subtasks} that must synthesize multiple fields into coherent explanations (e.g., component-level interpretation at L2, semantic analysis at L3, diachronic tracing at L4), LLM generation is necessary. However, general-purpose LLMs lack the specialized paleographic knowledge required for ACCE, and unconstrained generation often produces plausible-sounding but factually incorrect answers.

\begin{figure}
    \centering
    \includegraphics[width=1\linewidth]{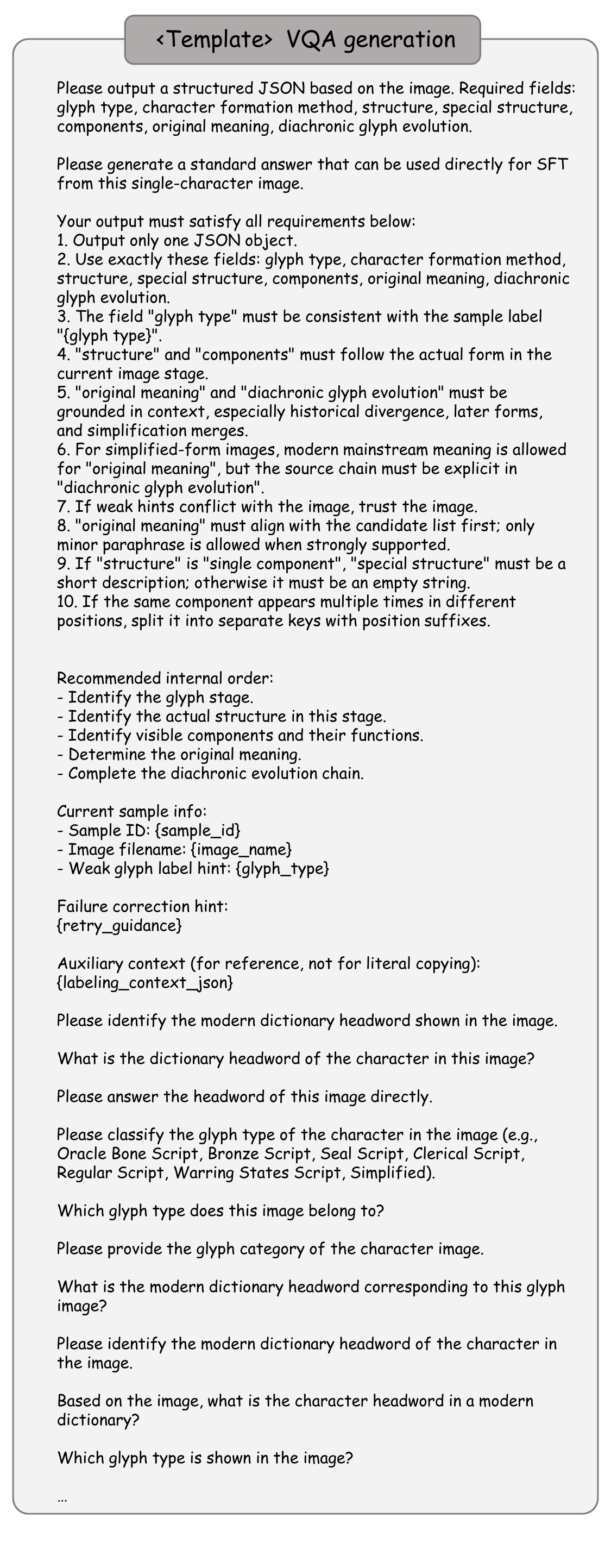}
    \vspace{-30pt}
    \caption{Prompt template for VQA generation. The prompt integrates to generate high-quality SFT data.}
    \label{fig:vqa_generation}
\end{figure}

Our templates address this by (1)~injecting the verified structured metadata as dynamic context, constraining the LLM to reason from the provided evidence rather than its parametric knowledge, and (2)~ guiding the model through the expert analytical workflow (identification $\to$ decomposition $\to$ interpretation $\to$ evolution tracing). Each template also includes multiple question phrasings per subtask to promote syntactic diversity. Fig.~\ref{fig:vqa_generation} presents a representative template.

\subsection{Quantitative Verification Details}
\label{sec:verification}

To make the quality-control process more transparent, we report the verification scope, sampling strategy, and revision statistics for each stage of our pipeline.

\textbf{Stage 1: Collection.}
For JieZi-Dataset, Stage~1 mainly involves digitization, including OCR correction and alignment of the digitized content with the original dictionary pages. Since the goal at this stage is only to ensure fidelity to the source dictionary rather than paleographic interpretation, the full scanned volume (over 2,000 pages) was checked by general annotators without requiring domain expertise. The main error types were OCR omissions and misrecognition of variant character forms. While errors were found in less than 1\% of entries, this full manual verification remains indispensable: as noted in the main paper, OCR errors disproportionately concentrate on rare and archaic glyphs, which are precisely the characters most critical to ACCE. Without exhaustive correction, even a small number of such errors would propagate into downstream structuring and QA generation, silently degrading the most informative entries in the dataset.

For JieZi-Bench, the source entries were also fully checked at the collection stage. Errors requiring correction were found in approximately 10\% of entries, a notably higher rate than that of JieZi-Dataset. This is expected for two reasons: (1)~JieZi-Bench integrates entries from multiple authoritative dictionaries, and cross-source alignment and deduplication introduce additional inconsistencies absent from the single-source Dataset; (2)~as a benchmark intended for evaluation, JieZi-Bench applies stricter acceptance criteria for entry completeness and accuracy, causing more entries to be flagged for revision. All identified errors were corrected before proceeding to the next stage.

\textbf{Stage 2: Structurization.}
At Stage~2, raw dictionary entries were converted into structured metadata records by an LLM-based extraction pipeline. Given the scale of JieZi-Dataset (over 500K entries), exhaustive manual verification was impractical at this stage. Instead, to verify dataset quality, we randomly sampled 10\% of the extracted metadata records. Experts inspected each sampled record by comparing it directly with the source text. Errors were found in approximately 1\% of the inspected subset.
The most common errors were missing structured fields and cross-source confusion. For example, the pipeline might misattribute a semantic explanation originally associated with the Oracle Bone form to the Bronze Inscription field of the same character.

For JieZi-Bench, all structured metadata records were exhaustively checked by domain experts. Errors requiring correction were found in approximately 5\% of entries; all identified errors were corrected before proceeding to VQA generation.

\begin{table}[t]
    \centering
    \caption{Training settings and hyper-parameters for different model scales.}
    \label{tab:training_details}
    \setlength{\tabcolsep}{5pt}
    \renewcommand{\arraystretch}{1.08}
    \begin{tabular}{lccc}
        \toprule
        \textbf{Setting} & \textbf{Qwen3.5-2B} & \textbf{Qwen3.5-4B} & \textbf{Qwen3.5-9B} \\
        \midrule
        Batch size & 128 & 128 & 64 \\
        Learning rate & 1e-5 & 1e-5 & 1e-5 \\
        LR schedule & cosine & cosine & cosine \\
        Warmup ratio & 0.05 & 0.05 & 0.05 \\
        Weight decay & 0.1 & 0.1 & 0.1 \\
        Optimizer & AdamW & AdamW & AdamW \\
        Max length & 2048 & 2048 & 1024 \\
        Epochs & 5 & 5 & 5 \\
        Precision & bf16 & bf16 & bf16 \\
        Hardware & 8 910B & 8 910B & 8 910B \\
        Training time & 96 NPU hours & 192 NPU hours & 384 NPU hours \\
        \bottomrule
    \end{tabular}
\end{table}

\textbf{Stage 3: VQA Generation and Verification.}
At Stage~3, VQA pairs were generated from the verified metadata by an LLM guided by expert-designed templates. The use of templates is essential because unconstrained LLM generation would suffer from three problems:
(1)~\textit{format inconsistency}, where question phrasing and answer granularity vary unpredictably, hindering standardized evaluation;
(2)~\textit{uneven subtask coverage}, where the LLM tends to favor subtasks it handles well while underrepresenting harder ones such as COME and EVOI;
(3)~\textit{hallucination risk}, where the LLM may fabricate etymological explanations or evolutionary paths absent from the source dictionaries.
Our template-guided approach addresses all three issues. Each template prescribes the question format and expected answer scope for a given subtask, ensuring uniform coverage and consistent granularity. The LLM then generates natural-language VQA pairs grounded in the metadata fields specified by the template, combining the scalability of LLM generation with the controllability of structured constraints.

Despite these safeguards, template-guided generation introduces its own errors. The generation process can produce image-content mismatches (e.g., pairing a glyph image with an answer describing a different script period), and natural-language answers may deviate from the source text in phrasing or scope. Moreover, VQA-level verification covers broader dimensions than Stage~2, including not only field-level correctness but also the naturalness of question-answer formulations and the appropriateness of paired images, which increases the likelihood of flagging issues.

For JieZi-Dataset, we randomly inspected 5,000 VQA instances. Since the underlying metadata had already been verified at Stage~2, the sampling ratio was reduced accordingly. Errors were found in approximately 3\% of the inspected subset; the slightly higher rate compared to Stage~2 reflects the new error sources and broader verification scope described above rather than upstream propagation.

For JieZi-Bench, all VQA pairs were manually checked, since benchmark reliability is critical for evaluation. Errors requiring correction were found in approximately 10\% of entries, consistent with the pattern observed at earlier stages: the benchmark's stricter acceptance criteria and cross-source complexity lead to a higher correction rate.

Overall, JieZi-Dataset adopts random spot-checking at Stages~2 and 3 to balance scale and quality, while JieZi-Bench uses exhaustive manual verification at all stages. These combined strategies ensure that JieZi-Bench achieves expert-level reliability for evaluation, while JieZi-Dataset maintains sufficient quality for training at scale.

\section{Experimental Details and Full Results}
\label{sec:exp_details}

\subsection{Additional Training Details}
\label{sec:training_details}

Table~\ref{tab:training_details} summarizes the hyperparameter settings for instruction tuning. All experiments were conducted on 8 Ascend 910B NPUs.

\subsection{Ablation Studies}
\label{sec:ablation}

Tab.~\ref{tab:ablation_data_stage} reports the impact of each data construction stage on Qwen3.5-2B. The first row corresponds to the few-shot base model without fine-tuning. We select CHAR, COMF, COMI, and EVOI as representative subtasks, as they span all four ACCE levels and cover both classification and generation objectives.

Each stage contributes a distinct gain profile. Structured metadata produces the largest improvement on the classification-oriented COMF (+16.0), since explicit component-function labels offer dense supervision that directly matches this subtask. In contrast, gains on generative subtasks remain moderate (COMI~+7.1, FAC~+7.5). VQA reformatting reverses this pattern: the generative subtasks benefit most (COMI~+13.0, FAC~+10.2), whereas COMF improves by only +1.7. This contrast suggests that recasting structured records as natural-language QA pairs trains the model to \emph{articulate} multi-step reasoning rather than simply retrieve labels.

Open-source augmentation, in turn, primarily strengthens visual robustness (CHAR~+6.5) and scholarly expression (SCE~+4.8). Factual subtasks show smaller gains (COMF~+1.3, FAC~+2.3), confirming that their performance is bottlenecked by knowledge rather than visual diversity. Taken together, the three stages address complementary dimensions of factual grounding, reasoning articulation, and visual robustness. Their consistent, non-overlapping improvements validate the necessity of each stage in the proposed pipeline.

\begin{table}[t]
\centering
\small
\setlength{\tabcolsep}{3.8pt}
\renewcommand{\arraystretch}{1.08}
\caption{Ablation study on different data construction stages.
\textbf{Struct.}: structured metadata; \textbf{VQA}: After VQA generation; \textbf{Open-src}: VQA with open-source data augmentation.}
\label{tab:ablation_data_stage}
\resizebox{\columnwidth}{!}{
\begin{tabular}{ccc ccccc}
\toprule
\textbf{Struct.} & \textbf{VQA} & \textbf{Open-src} & \textbf{CHAR} $\uparrow$ & \textbf{COMF} $\uparrow$ & \textbf{COMI} $\uparrow$ & \multicolumn{2}{c}{\textbf{EVOI}} \\
\cmidrule(lr){7-8}
 & & & & & & \textbf{FAC} $\uparrow$ & \textbf{SCE} $\uparrow$ \\
\midrule
           &            &            &  18.9  &  13.7  &  17.0  &   7.9  &  20.2  \\
\cmark     &            &            &  27.0  &  29.7  &  24.1  &  15.4  &  26.3  \\
\cmark     & \cmark     &            &  35.3  &  31.4  &  37.1  &  25.6  &  31.9  \\
\cmark     & \cmark     & \cmark     &  41.8  &  32.7  &  41.3  &  27.9  &  36.7  \\
\bottomrule
\end{tabular}
}
\end{table}

\subsection{Human Validation of the LLM-as-a-Judge}
\label{sec:human_agreement}

To rigorously validate the reliability of our LLM-as-a-Judge protocol for the Evolution Interpretation (EVOI) task, we conducted a human-LLM agreement study. We randomly sampled 200 responses in total across 4 representative models to ensure a diverse distribution of response qualities. Three human experts with backgrounds in Chinese paleography independently evaluated these responses on a scale of 1 to 5, focusing on two dimensions: Fact Alignment (FAC) and Scholarly Expression (SCE). 

We computed the Pearson ($r$) and Spearman ($\rho$) correlation coefficients between the automated LLM judge scores and the human ground truth. 
As shown in Tab.~\ref{tab:llm_judge_agreement}, the LLM judge exhibits strong correlations ($> 0.60$) with expert evaluations across both metrics. The results confirm that the LLM judge aligns closely with human scholarly standards, demonstrating its effectiveness and reliability as an automated evaluation metric for ancient character exegesis.

\textbf{BERTScore validation.}
We further validate BERTScore, used for the open-ended ORIM and COMI tasks, under the same expert protocol. Three paleography experts scored 200 responses from four models on a 1--5 scale, and we computed Pearson and Spearman correlations between BERTScore and averaged human ratings.

\begin{table}[t]
    \centering
    \caption{Correlation between human expert ratings and BERTScore on the ORIM and COMI tasks.}
    \label{tab:bertscore_agreement}
    \begin{tabular}{lcc}
        \toprule
        \textbf{Task} & \textbf{Pearson ($r$)} & \textbf{Spearman ($\rho$)} \\
        \midrule
        ORIM & 0.71 & 0.71 \\
        COMI & 0.73 & 0.72 \\
        \bottomrule
    \end{tabular}
\end{table}

As shown in Tab.~\ref{tab:bertscore_agreement}, BERTScore achieves strong human correlation (Pearson $\geq 0.71$) on both tasks. Together with the LLM-as-a-Judge validation in Tab.~\ref{tab:llm_judge_agreement}, these results confirm that both automated metrics provide sufficient reliability for comparative evaluation of open-ended exegetical responses.

\begin{figure}[t]
    \centering
    \includegraphics[width=1\linewidth]{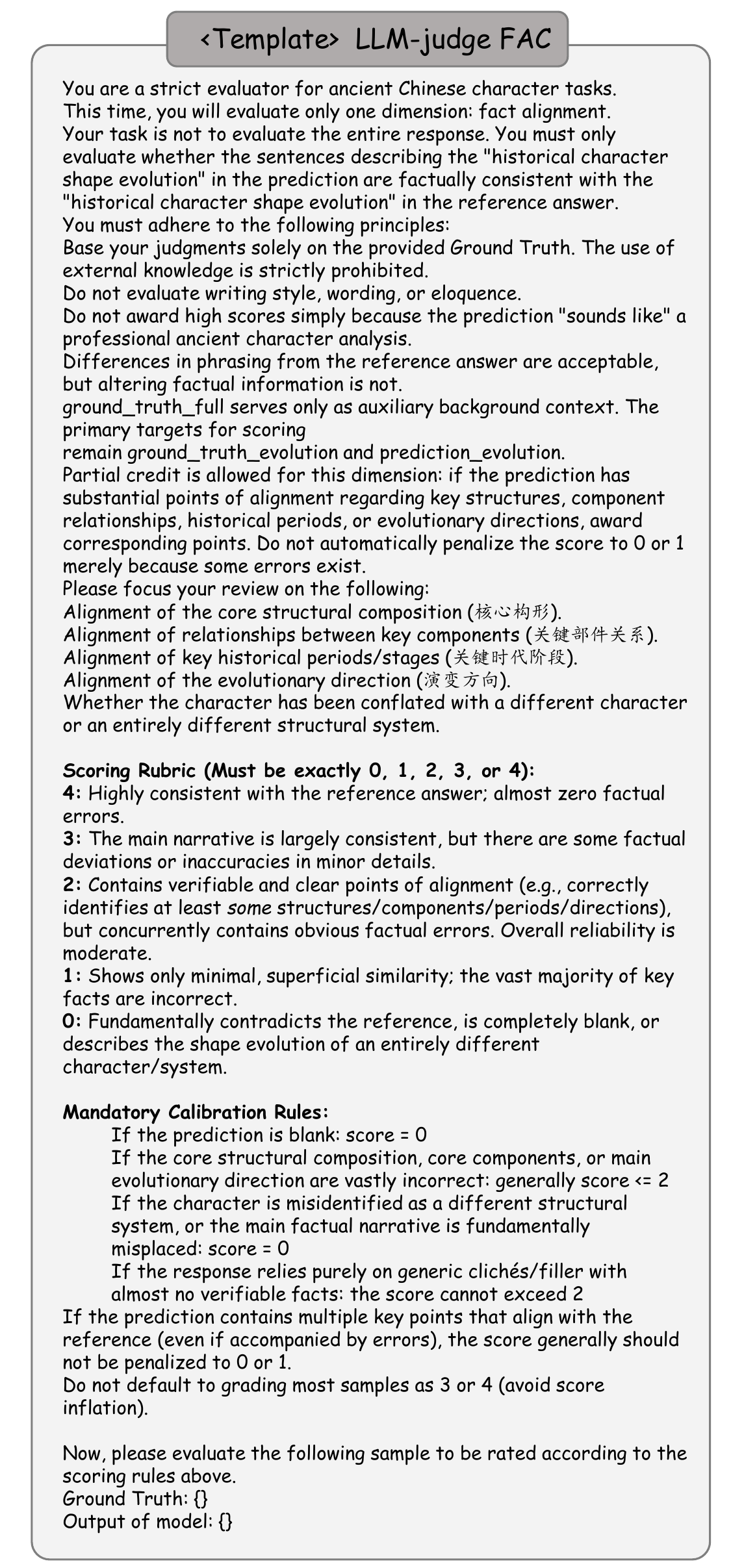}
    \caption{Prompt template for evaluating fact alignment within our LLM-as-a-judge protocol.}
    \label{fig:fac-prompt}
\end{figure}

\begin{figure}[t]
    \centering
    \includegraphics[width=1\linewidth]{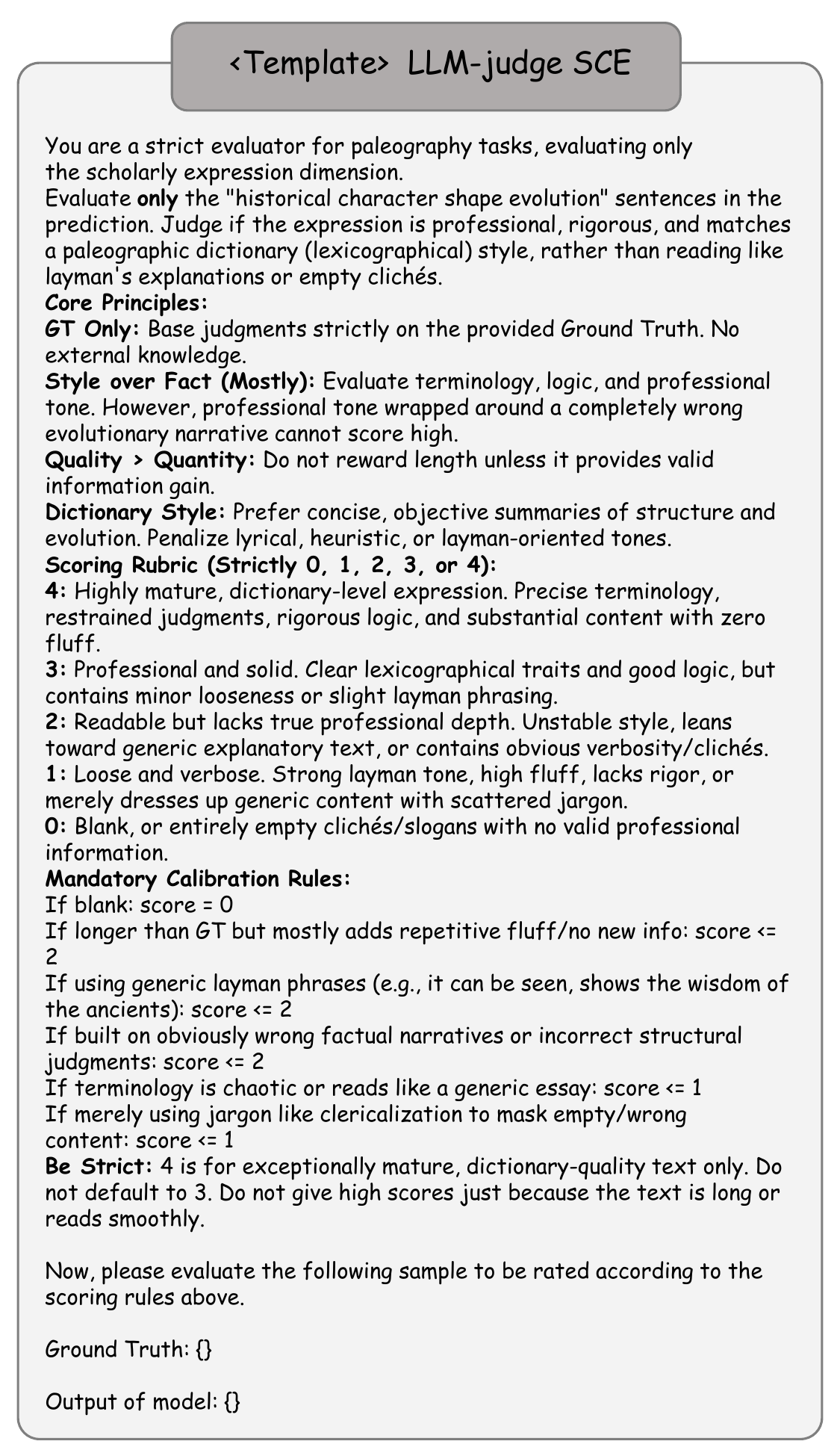}
    \caption{Prompt template for evaluating scholarly expression within our LLM-as-a-judge protocol.}
    \label{fig:sce-prompt}
\end{figure}

\textbf{Judging prompts.}
We provide the full prompts used in our LLM-based judging protocol for EVOI. Fig.~\ref{fig:fac-prompt} presents the prompt for evaluating \textbf{Fact Alignment}, while Fig.~\ref{fig:sce-prompt} presents the prompt for evaluating \textbf{Scholarly Expression}. We used Doubao-seed-2-0-lite-260215~\cite{bytedance_seed_2_0_launch_2026} as our LLM-as-a-judge model.

\begin{table}[t]
    \centering
    \caption{Correlation between human expert ratings and LLM judge scores on the EVOI task.}
    \label{tab:llm_judge_agreement}
    \begin{tabular}{lcc}
        \toprule
        \textbf{Metric} & \textbf{Pearson ($r$)} & \textbf{Spearman ($\rho$)} \\
        \midrule
        FAC  & 0.68 & 0.71 \\
        SCE  & 0.61 & 0.64 \\
        \bottomrule
    \end{tabular}
\end{table}

\subsection{Main Results on JieZi-Bench and Analysis}
\label{sec:main_results}

\begin{table*}[t]
\centering
\small
 \setlength{\tabcolsep}{4.2pt} 
\renewcommand{\arraystretch}{1.12}
\caption{Main results on JieZi-Bench across four paleographic levels: \textbf{L1} (Basic Info: CHAR, SCRC), \textbf{L2} (Glyph Form: STRC, COMR, COMF, COMI, FORC), \textbf{L3} (Meaning: ORIM), and \textbf{L4} (Evolution: COME, EVOI-FAC/SCE). All metrics are scaled to 0-100.}
\label{tab:main_results_supp}
\resizebox{\textwidth}{!}{
\begin{tabular}{l cccccccc c cc}
\toprule
\multirow{3}{*}{\textbf{Method}}
& \multicolumn{2}{c}{\textbf{L1}}
& \multicolumn{5}{c}{\textbf{L2}}
& \multicolumn{1}{c}{\textbf{L3}}
& \multicolumn{3}{c}{\textbf{L4}} \\
\cmidrule(lr){2-3} \cmidrule(lr){4-8} \cmidrule(lr){9-9} \cmidrule(lr){10-12}
& \multirow{2}{*}{\textbf{CHAR}$\uparrow$} & \multirow{2}{*}{\textbf{SCRC}$\uparrow$}
& \multirow{2}{*}{\textbf{STRC}$\uparrow$} & \multirow{2}{*}{\textbf{COMR}$\uparrow$} & \multirow{2}{*}{\textbf{COMF}$\uparrow$} & \multirow{2}{*}{\textbf{COMI}$\uparrow$} & \multirow{2}{*}{\textbf{FORC}$\uparrow$}
& \multirow{2}{*}{\textbf{ORIM}$\uparrow$}
& \multirow{2}{*}{\textbf{COME}$\uparrow$}
& \multicolumn{2}{c}{\textbf{EVOI}} \\
\cmidrule(lr){11-12}
& & & & & & & & & & \textbf{FAC}$\uparrow$ & \textbf{SCE}$\uparrow$ \\
\midrule
\multicolumn{12}{l}{\textit{Closed-source MLLMs}} \\
GPT-5.4 Thinking~\cite{gpt54thinking_system_card}
& 10.4 & 39.2
& 50.4 & 13.8 & 11.9 & 10.3 & 33.1
& 59.9
& 12.4 & 10.2 & 16.8 \\
Gemini 3.1 Pro~\cite{gemini31pro_model_card}
& 29.9 & 76.6
& 65.7 & 31.9 & 29.9 & 24.8 & 56.1
& 62.8
& 30.3 & 22.3 & 27.1 \\
Gemini 3.1 Flash~\cite{gemini31flashlite_model_card}
& 25.7 & 56.5
& 53.5 & 24.6 & 22.5 & 19.1 & 44.8
& 54.1
& 23.2 & 17.5 & 23.1 \\
Claude opus 4.6~\cite{claude_opus46_system_card}
& 22.9 & 65.8
& 63.5 & 28.0 & 26.3 & 22.3 & 50.6
& 59.3
& 26.7 & 19.8 & 26.3 \\
Doubao-Seed-2.0-pro~\cite{seed20_model_card}
& 36.6 & 68.5
& 69.8 & 37.8 & 34.4 & 29.9 & 59.9
& 63.2
& 36.1 & 31.7 & 38.7 \\
\midrule
\multicolumn{12}{l}{\textit{Open-source MLLMs}} \\
Kimi-K2.5~\cite{kimiteam2026kimik25visualagentic} (170B)
& 31.5 & 70.1
& 68.8 & 39.2 & 36.6 & 31.5 & 58.1
& 66.6
& 36.4 & 28.3 & 34.6 \\
GLM-4.6V~\cite{glm46v_paper} (107B)
& 18.2 & 49.7
& 48.9 & 21.8 & 20.4 & 18.2 & 42.2
& 48.2
& 20.9 & 15.5 & 20.2 \\
Qwen3.5-397b-a17b~\cite{qwen3.5}
& 26.2 & 67.1
& 69.4 & 32.1 & 29.9 & 26.5 & 54.4
& 59.5
& 28.3 & 22.2 & 30.7 \\
Qwen3.5-35b-a3b~\cite{qwen3.5}
& 26.6 & 61.4
& 66.5 & 32.4 & 29.3 & 27.3 & 57.8
& 65.2
& 31.0 & 19.6 & 25.6  \\
InternVL3.5-8B~\cite{wang2025internvl3_5}
& 00.1 & 09.0
& 14.6 & 0.0 & 0.0 & 0.0 & 21.4
& 58.0
& 0.0 & 24.8 & 0.1  \\
InternVL3.5-4B~\cite{wang2025internvl3_5}
& 0.0 & 0.0
& 0.9 & 0.0 & 0.0 & 0.0 & 8.2
& 24.3
& 0.0 & 0.0 & 0.0  \\
tonggu-vl-2B~\cite{cao2025tonggu}
& 3.1 & 05.9
& 0.2 & 0.0 & 0.0 & 0.0 & 4.6
& 12.9
& 0.0  & 1.1 & 0.2 \\
\midrule
Qwen3.5-2B~\cite{qwen3.5}
& 18.9 & 22.6
& 31.6 & 17.3 & 13.7 & 17.0 & 30.7
& 62.3
& 12.9 & 7.9 & 20.2 \\
\textbf{Qwen3.5-2B + JieZi-Dataset}
& 41.8\up{22.9} & 61.7\up{39.1}
& 72.7\up{41.1} & 42.6\up{25.3} & 32.7\up{19.0} & 41.3\up{24.3} & 63.3\up{32.6}
& 65.7\up{3.4}
& 28.3\up{15.4} & 27.9\up{20.0} & 36.7\up{16.5} \\
\midrule
Qwen3.5-4B~\cite{qwen3.5}
& 3.6 & 6.6
& 1.0 & 4.5 & 3.9 & 4.3 & 5.9
& 6.5
& 3.6 & 2.0 & 3.6 \\
\textbf{Qwen3.5-4B + JieZi-Dataset}
& 49.0\up{45.4} & 67.6\up{61.0}
& 74.1\up{73.1} & 47.2\up{42.7} & 37.9\up{34.0} & 42.4\up{38.1} & 64.8\up{58.9}
& 65.9\up{59.4}
& 33.2\up{29.6} & 28.5\up{26.5} & 37.2\up{33.6} \\
\midrule
Qwen3.5-9B~\cite{qwen3.5}
& 22.3 & 51.0
& 40.2 & 27.4 & 24.6 & 26.1 & 49.1
& 64.7
& 22.4 & 15.0 & 23.5 \\
\textbf{Qwen3.5-9B + JieZi-Dataset}
& \textbf{52.6}\up{30.3} & \textbf{80.1}\up{29.1}
& \textbf{74.7}\up{34.5} & \textbf{48.1}\up{20.7} & \textbf{38.4}\up{13.8} & \textbf{45.7}\up{19.6} & \textbf{66.5}\up{17.4}
& \textbf{66.7}\up{2.0}
& \textbf{45.4}\up{23.0} & \textbf{32.5}\up{17.5} & \textbf{40.6}\up{17.1} \\
\bottomrule
\end{tabular}
}
\end{table*}

\begin{figure}
    \centering
    \includegraphics[width=1\linewidth]{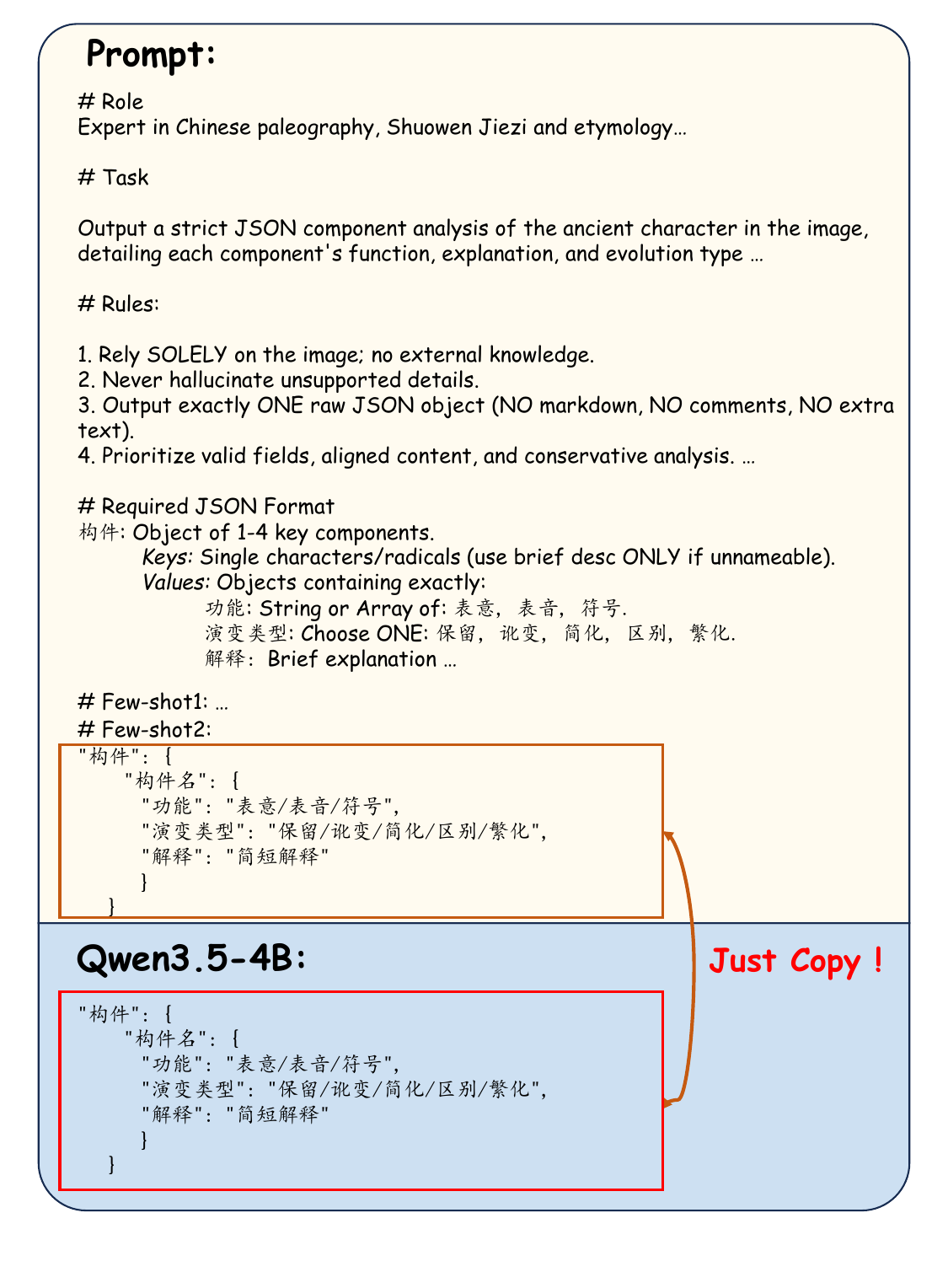}
    \caption{An example of anomalous output from few-shot Qwen3.5-4B, illustrating the copying bias.}
    \label{fig:qwen-badcase}
\end{figure}

Due to space limitations in the main text, we present the complete experimental results in Table~\ref{tab:main_results_supp}, including additional models not shown in the main paper. Below, we provide a detailed analysis of the key findings.

\textbf{Scaling law holds within model families.}
Tab.~\ref{tab:main_results_supp} includes additional model variants not reported in the main paper, enabling within-family comparisons among general MLLMs. Across all families, larger variants consistently outperform smaller ones. Gemini~3.1~Pro surpasses Gemini~3.1~Flash on every metric (e.g., CHAR: 29.9 vs.\ 25.7; SCRC: 76.6 vs.\ 56.5), Qwen3.5-397b-a17b outperforms 35b-a3b (e.g., SCRC: 67.1 vs.\ 61.4), and InternVL3.5-8B consistently exceeds its 4B counterpart (e.g., ORIM: 58.0 vs.\ 24.3). These within-family comparisons, together with the cross-family results in the main paper, confirm that the general scaling trend holds for ACCE.

\textbf{Domain-specialized pretraining does not guarantee ACCE proficiency.}
tonggu-vl-2B~\cite{cao2025tonggu} is a model specifically fine-tuned on massive ancient Chinese document corpora for tasks such as text recognition and reading comprehension. Despite this domain-specialized pretraining, it scores below 5\% on most ACCE subtasks (e.g., CHAR: 3.1, STRC: 0.2, COMR/COMF/COMI: 0.0) and achieves only 12.9 on ORIM. This reveals a critical distinction: \emph{recognizing and reading ancient texts} is fundamentally different from \emph{performing scholarly exegesis}, which requires structured reasoning about glyph form, component function, semantic origin, and diachronic evolution. The failure of tonggu-vl demonstrates that ACCE poses genuinely novel challenges beyond the reach of existing domain-specific models, thereby validating both the task formulation and the necessity of JieZi-Dataset as a dedicated training resource. Conversely, once equipped with JieZi-Dataset, even a 9B model outperforms general MLLMs that are one to two orders of magnitude larger (e.g., Kimi-K2.5 170B, GLM-4.6V 107B, Qwen3.5-397b) on nearly every subtask, further confirming that high-quality domain-specific data is a far more effective lever than model scale alone for knowledge-intensive tasks like ACCE.

\textbf{Remark on the anomalous behavior of Qwen3.5-4B.}
As shown in Tab.~\ref{tab:main_results_supp}, few-shot Qwen3.5-4B scores extremely low across all eleven ACCE subtasks (e.g., CHAR: 3.6, STRC: 1.0, ORIM: 6.5), performing substantially below both the smaller 2B and the larger 9B variants. We verified that all three models share identical inference pipelines, prompts, and decoding parameters, ruling out implementation errors. Manual inspection of the 4B outputs (Fig.~\ref{fig:qwen-badcase}) reveals that the model tends to reproduce content from the few-shot demonstrations rather than generating answers grounded in the current query, producing outputs that bear no meaningful correspondence to the target glyph.

Since Qwen3.5's training details are not publicly disclosed, we can only hypothesize about the root cause based on observed behavior. We identify two plausible contributing factors.
(1)~\emph{Copying bias in few-shot in-context learning (ICL)}: The model reproduces demonstration content instead of reasoning about the current query. This behavior directly aligns with prior findings that LLMs can develop a copying bias under in-context learning, reproducing surface patterns from examples rather than inducing the underlying task~\cite{olsson2022icl,min2022rethinking}.
(2)~\emph{U-shaped scaling}: The 2B (functional) $\to$ 4B (collapsed) $\to$ 9B (functional) pattern is consistent with the U-shaped scaling phenomenon, where medium-sized models are drawn toward easier competing behaviors instead of the target task~\cite{wei2023inverse,mckenzie2023inverse}.

After fine-tuning on JieZi-Dataset, Qwen3.5-4B fully recovers to the expected scaling order, achieving the largest absolute improvements among the three model scales (e.g., STRC: +73.1, ORIM: +59.4 vs.\ 2B's +41.1 for STRC and +3.4 for ORIM, and 9B's +34.5 for STRC and +2.0 for ORIM). This confirms that the bottleneck is not model capacity but misaligned few-shot behavior, which domain-specific supervision effectively overrides~\cite{zeng2024scaling}.

\subsection{Baseline Comparison}
\label{sec:baseline}

To evaluate whether the performance gains from JieZi-Dataset stem from domain-specific reasoning rather than mere knowledge exposure, we compare supervised fine-tuning (SFT) against three alternative training-free strategies on Qwen3.5-9B: (1)~few-shot in-context learning (ICL) with top-3 retrieved examples, (2)~retrieval-augmented generation (RAG), and (3)~dictionary-context prompting. All baselines use the same model backbone, JieZi-Bench split, decoding settings, and evaluation metrics as Tab.~\ref{tab:main_results_supp}. In ACCE, questions are generic templates that carry no character-specific information; the character identity is conveyed entirely by the glyph image, making a no-image baseline inapplicable to this task.

For few-shot ICL, we retrieve the top-3 training examples using only the query glyph image, without using benchmark labels or reference answers. For RAG, the model receives top-3 retrieved training metadata and QA records as context but is not fine-tuned. Dictionary-context prompting provides the retrieved dictionary-style context directly in the prompt. For both few-shot ICL and RAG, we use SigLIP2-SO400M-patch14-384 as the image retriever: the query glyph and training glyphs are encoded into visual embeddings, and the top-3 nearest training examples are selected by cosine similarity.

\begin{table*}[t]
\centering
\small
\setlength{\tabcolsep}{4.2pt}
\renewcommand{\arraystretch}{1.12}
\caption{Comparison of SFT on JieZi-Dataset against alternative training-free strategies on Qwen3.5-9B, evaluated on JieZi-Bench. All methods use the same backbone, decoding settings, and evaluation metrics.}
\label{tab:baseline_comparison}
\resizebox{\textwidth}{!}{
\begin{tabular}{l cccccccc c cc}
\toprule
\multirow{2}{*}{\textbf{Method}}
& \multicolumn{2}{c}{\textbf{L1}}
& \multicolumn{5}{c}{\textbf{L2}}
& \multicolumn{1}{c}{\textbf{L3}}
& \multicolumn{3}{c}{\textbf{L4}} \\
\cmidrule(lr){2-3} \cmidrule(lr){4-8} \cmidrule(lr){9-9} \cmidrule(lr){10-12}
& \textbf{CHAR}$\uparrow$ & \textbf{SCRC}$\uparrow$
& \textbf{STRC}$\uparrow$ & \textbf{COMR}$\uparrow$ & \textbf{COMF}$\uparrow$ & \textbf{COMI}$\uparrow$ & \textbf{FORC}$\uparrow$
& \textbf{ORIM}$\uparrow$
& \textbf{COME}$\uparrow$ & \textbf{FAC}$\uparrow$ & \textbf{SCE}$\uparrow$ \\
\midrule
Qwen3.5-9B & 22.3 & 51.0 & 40.2 & 27.4 & 24.6 & 26.1 & 49.1 & 64.7 & 22.4 & 15.0 & 23.5 \\
+ Few-shot ICL (top-3) & 21.4 & 56.7 & 46.9 & 27.6 & 24.5 & 18.8 & 51.5 & 53.8 & 26.1 & 14.5 & 19.3 \\
+ RAG & 30.3 & 50.9 & 34.2 & 31.8 & 27.8 & 25.2 & 58.3 & 57.5 & 30.3 & 19.2 & 21.8 \\
+ Dict-context prompting & 22.2 & 37.0 & 49.9 & 27.4 & 21.4 & 18.9 & 58.3 & 53.8 & 25.9 & 15.6 & 18.3 \\
+ JieZi-Dataset (SFT) & \textbf{52.6} & \textbf{80.1} & \textbf{74.7} & \textbf{48.1} & \textbf{38.4} & \textbf{45.7} & \textbf{66.5} & \textbf{66.7} & \textbf{45.4} & \textbf{32.5} & \textbf{40.6} \\
\bottomrule
\end{tabular}
}
\end{table*}

As shown in Tab.~\ref{tab:baseline_comparison}, SFT on JieZi-Dataset outperforms all alternatives across all subtasks. Few-shot ICL with retrieved examples provides substantial improvement over the base model, indicating that in-context paleographic examples are informative. However, RAG and dictionary-context prompting, which inject the same paleographic knowledge in different forms, still fall far short of SFT. This confirms that the performance gain stems from learning domain-specific reasoning patterns through supervised fine-tuning, not merely from exposure to paleographic knowledge in any form.

\subsection{Data Scaling Analysis}
\label{sec:scaling}

To examine whether the full 500K-scale training set is necessary or whether comparable performance is achievable with fewer QA pairs, we train Qwen3.5-2B with nested 25\%, 50\%, 75\%, and 100\% subsets of JieZi-Dataset. The subsets are strictly nested: the 25\% subset is contained in the 50\% subset, which is contained in the 75\% subset. All runs use the same hyperparameters as Tab.~\ref{tab:training_details} and are evaluated on the unchanged JieZi-Bench.

\begin{table*}[t]
\centering
\small
\setlength{\tabcolsep}{4.2pt}
\renewcommand{\arraystretch}{1.12}
\caption{Data scaling analysis on Qwen3.5-2B with nested subsets of JieZi-Dataset, evaluated on JieZi-Bench.}
\label{tab:data_scaling}
\resizebox{\textwidth}{!}{
\begin{tabular}{l cccccccc c cc}
\toprule
\multirow{2}{*}{\textbf{Training Data}}
& \multicolumn{2}{c}{\textbf{L1}}
& \multicolumn{5}{c}{\textbf{L2}}
& \multicolumn{1}{c}{\textbf{L3}}
& \multicolumn{3}{c}{\textbf{L4}} \\
\cmidrule(lr){2-3} \cmidrule(lr){4-8} \cmidrule(lr){9-9} \cmidrule(lr){10-12}
& \textbf{CHAR}$\uparrow$ & \textbf{SCRC}$\uparrow$
& \textbf{STRC}$\uparrow$ & \textbf{COMR}$\uparrow$ & \textbf{COMF}$\uparrow$ & \textbf{COMI}$\uparrow$ & \textbf{FORC}$\uparrow$
& \textbf{ORIM}$\uparrow$
& \textbf{COME}$\uparrow$ & \textbf{FAC}$\uparrow$ & \textbf{SCE}$\uparrow$ \\
\midrule
Qwen3.5-2B (base) & 18.9 & 22.6 & 31.6 & 17.3 & 13.7 & 17.0 & 30.7 & 62.3 & 12.9 & 7.9 & 20.2 \\
+ 25\% JieZi-Dataset & 34.0 & 33.3 & 67.7 & 36.5 & 30.5 & 25.5 & 45.0 & 56.1 & 23.0 & 21.1 & 28.5 \\
+ 50\% JieZi-Dataset & 37.6 & 43.5 & 66.4 & 37.7 & 30.6 & 26.0 & 60.4 & 56.5 & 23.9 & 23.5 & 30.5 \\
+ 75\% JieZi-Dataset & 40.6 & 52.1 & 67.0 & 40.1 & 31.4 & 28.1 & 62.3 & 59.9 & 28.0 & 25.8 & 33.2 \\
+ 100\% JieZi-Dataset & \textbf{41.8} & \textbf{61.7} & \textbf{72.7} & \textbf{42.6} & \textbf{32.7} & \textbf{41.3} & \textbf{63.3} & \textbf{65.7} & \textbf{28.3} & \textbf{27.9} & \textbf{36.7} \\
\bottomrule
\end{tabular}
}
\end{table*}

As shown in Tab.~\ref{tab:data_scaling}, the macro average increases monotonically from 36.5 at 25\% to 46.8 at 100\% (+10.3). While 25\% already yields substantial gains over the base model (+13.3), the full set still brings clear improvements on most subtasks (e.g., SCRC 33.3$\to$61.7, COMI 25.5$\to$41.3), confirming that the full 500K scale provides real marginal benefit rather than redundant volume.

\subsection{Paraphrase Robustness Test}
\label{sec:paraphrase}

To test whether the fine-tuned model overfits to fixed template question patterns rather than learning genuine exegetical ability, we conduct a controlled paraphrase robustness test. For each instance across all subtasks, we replace the original question with two independently paraphrased versions while keeping the glyph image, reference answer, model (Qwen3.5-2B fine-tuned on JieZi-Dataset), decoding settings, and evaluation metrics unchanged. This isolates the effect of question wording from all other factors. The two paraphrases are generated by prompting Gemini to rewrite each template question with different lexical choices and sentence structures while preserving the query intent.

\begin{table*}[t]
\centering
\small
\setlength{\tabcolsep}{4.2pt}
\renewcommand{\arraystretch}{1.12}
\caption{Paraphrase robustness test on Qwen3.5-2B fine-tuned with JieZi-Dataset. ``Original'' uses the standard template questions; ``Prompt~2'' and ``Prompt~3'' are independently paraphrased versions. Avg.~$\Delta$ reports the mean score change across the two paraphrased variants relative to the original.}
\label{tab:paraphrase}
\resizebox{\textwidth}{!}{
\begin{tabular}{l cccccccc c cc}
\toprule
\multirow{2}{*}{\textbf{Question}}
& \multicolumn{2}{c}{\textbf{L1}}
& \multicolumn{5}{c}{\textbf{L2}}
& \multicolumn{1}{c}{\textbf{L3}}
& \multicolumn{3}{c}{\textbf{L4}} \\
\cmidrule(lr){2-3} \cmidrule(lr){4-8} \cmidrule(lr){9-9} \cmidrule(lr){10-12}
& \textbf{CHAR}$\uparrow$ & \textbf{SCRC}$\uparrow$
& \textbf{STRC}$\uparrow$ & \textbf{COMR}$\uparrow$ & \textbf{COMF}$\uparrow$ & \textbf{COMI}$\uparrow$ & \textbf{FORC}$\uparrow$
& \textbf{ORIM}$\uparrow$
& \textbf{COME}$\uparrow$ & \textbf{FAC}$\uparrow$ & \textbf{SCE}$\uparrow$ \\
\midrule
Original & 41.8 & 61.7 & 72.7 & 42.6 & 32.7 & 41.3 & 63.3 & 65.7 & 28.3 & 27.9 & 36.7 \\
Prompt 2 & 44.6 & 58.0 & 63.9 & 42.5 & 33.0 & 39.6 & 64.2 & 57.3 & 41.1 & 25.8 & 29.8 \\
Prompt 3 & 42.6 & 59.8 & 67.6 & 34.6 & 31.6 & 33.7 & 56.8 & 59.9 & 33.2 & 21.0 & 44.5 \\
\midrule
Avg.~$\Delta$ & +1.8 & $-$2.8 & $-$7.0 & $-$4.1 & $-$0.4 & $-$4.7 & $-$2.8 & $-$7.1 & +8.9 & $-$4.5 & +0.5 \\
\bottomrule
\end{tabular}
}
\end{table*}

As shown in Tab.~\ref{tab:paraphrase}, the fine-tuned model remains stable across paraphrased questions. Although some subtasks show moderate drops (e.g., STRC $-$7.0, ORIM $-$7.1), the overall performance remains substantially above the base model across all variants. This indicates that the performance gains are not mainly driven by surface-level template pattern matching but by learned glyph-grounded exegetical content.

\end{document}